\documentclass[sigconf,9pt]{acmart}
\setcopyright{none}
\renewcommand\footnotetextcopyrightpermission[1]{}
\setcopyright{none}
\makeatletter
\renewcommand\@formatdoi[1]{\ignorespaces}
\makeatother

\AtBeginDocument{%
  \providecommand\BibTeX{{%
    \normalfont B\kern-0.5em{\scshape i\kern-0.25em b}\kern-0.8em\TeX}}}

\usepackage{algpseudocode}
\usepackage{subfig}
\usepackage{float}
\usepackage{enumitem}
\usepackage{algorithm2e}

\begin{document}

\title{PreMa: Predictive Maintenance of Solenoid Valve in Real-Time at Embedded Edge-Level}

\author{Prajwal BN}
\affiliation{%
   \institution{Indian Institute of Science}
   \city{Bengaluru}
   \country{India}
   }
\email{prajwalbn@gmail.com}

\author{Harsha Yelchuri}
\affiliation{%
   \institution{Indian Institute of Science}
   \city{Bengaluru}
   \country{India}
   }
\email{harshay@iisc.ac.in}

\author{Vishwanath Shastry}
\affiliation{%
   \institution{Indian Institute of Science}
   \city{Bengaluru}
   \country{India}
   }
\email{vshastry816@gmail.com }

\author{T. V. Prabhakar}
\affiliation{%
   \institution{Indian Institute of Science}
   \city{Bengaluru}
   \country{India}
   }
\email{tvprabs@iisc.ac.in}





\begin{abstract}
  In industrial process automation, sensors (pressure, temperature, etc.), controllers, and actuators (solenoid valves, electro-mechanical relays, circuit breakers, motors, etc.) make sure that production lines are working under the pre-defined conditions. When these systems malfunction or sometimes completely fail, alerts have to be generated in real-time to make sure not only production quality is not compromised but also safety of humans and equipment is assured. In this work, we describe the construction of a smart and real-time edge-based electronic product called PreMa, which is basically a sensor for monitoring the health of a Solenoid Valve (SV).  PreMa is compact, low power, easy to install, and cost effective.  It has data fidelity and measurement accuracy comparable to signals captured using high end equipment. The smart solenoid sensor runs TinyML, a compact version of TensorFlow (a.k.a. TFLite) machine learning framework. While fault detection inferencing is in-situ, model training uses mobile phones to accomplish the `on-device' training. Our product evaluation shows that the sensor is able to differentiate between the distinct types of faults.  These faults include: (a) Spool stuck (b) Spring failure and (c) Under voltage. Furthermore, the product provides maintenance personnel, the remaining useful life (RUL) of the SV.  The RUL provides assistance to decide valve replacement or otherwise. We perform an extensive evaluation on optimizing metrics related to performance of the entire system (i.e. embedded platform and the neural network model). The proposed implementation is such that, given any electro-mechanical actuator with similar transient response to that of the SV, the system is capable of condition monitoring, hence presenting a first of its kind generic infrastructure.
\end{abstract}

\ccsdesc[500]{Computer systems organization~Sensors and actuators}

\keywords{Fault Detection, Edge Intelligence, RUL, Solenoid Valve, TinyML}

\maketitle

\pagestyle{plain}
\section{Introduction and Motivation\label{section:intro}}
The SV finds a broad spectrum of applications in industrial automation, automobiles, agriculture, medicine, aerospace, railways and in many other domains. For e.g., in the industrial sector, it finds applications to facilitate switching operations, pilot plant control loops, and in a plethora of original equipment manufacturing applications \cite{[33]}. Whereas in the agricultural field, it is deployed for the automation of irrigation systems \cite{[36]}. In the automobile sector, it is used in coolant regulator and for the automation of water fuel separation systems. The medical industry employs SVs in dialysis systems, drug flow regulators, clinical processing plants and ventilators \cite{[31]}. To sum up, SV functions as an electro$-$mechanical actuator in various engineering systems.

Over time, the efficiency of these valves might deteriorate, thus compromising the entire process for which it is employed. These types of compromises due to the quality issues over time can lead to a disastrous phenomenon while being engaged in crucial processes. For instance, the SV used in a medical-grade ventilator, is responsible for several of the ventilator's critical functions, such as actuating the mixing cylinder (air and oxygen) and the delivery valve. Many scenarios can be speculated if one of these SVs deployed in the ventilator malfunctions, which can lead to breathlessness and also prove fatal. This situation can be circumvented by real-time predictive maintenance of such crucial components. One such economical and accurate predictive maintenance is the estimation of RUL and fault detection of SV.  

Witnessing the extensive applications of SV in the areas mentioned above, in order to monitor the health of SV, it is vital to understand the factors leading to the deterioration in its performance and eventually its failure. Identifying various faults \cite{[12]} that could occur in the SV during its lifetime is one of the main challenges faced in diagnosing its health. The leading causes of fault in the valves depend not only on its intrinsic component's degradation , but also on external factors. Problems arising due to intrinsic components can include spool stuck caused due to deposition of particles, spring failures, and issues with coil due to overheating \cite{[5]}. External factors can include unregulated power supply, the pressure difference across the ports, the temperature of the SV, and surrounding environmental conditions. 

The convergence of embedded systems and machine learning is highlighted by TinyML \cite{[29]}. This technology allows researchers to create unique applications in the combined area of embedded systems and machine learning tailored to their needs, hence mitigating the requirement for a cloud-based aggregator to act as an intelligent action provider. Collectively, it is a standalone cognitive solution providing low power, low latency, narrow bandwidth and data privacy \cite{[28]}.

The incessant growth in the field of TinyML and the importance of monitoring the health of the SV across its application spectrum has routed us to devise a solution that facilitates real-time edge-device based predictive maintenance of SV in its industrial operation. Precisely, our system extracts required features from the transient response of the SV and concurrently diagnoses it by reporting its working condition and remaining useful life with the help of neural network (NN) models. Unlike the traditional approaches of training the model in offline mode, we present a novel way of on-device training on mobile phones. This unconventional process of training enables refactoring of the model, making it adaptive to sector-specific needs. The proposed on-device system is characterized by real-time feature extraction, training, and inference, aiding deployment with minimal latency. 

To keep the cost of the deployment economical, our system is deployed on edge platforms and targets cost-effective sensors to artifact the traditional high end expensive setup. Our proposed system is generic and can be applied to similar types of actuators like electromagnetic relays, circuit breakers, and other distinct types of valves.

The following are our noteworthy contributions$:$
\begin{itemize}
\item Implemented an end-to-end edge device application (TinyML based) for real-time predictive maintenance (Fault Detection and RUL) of SV.
\item A custom built, intelligent electronic product that encapsulates data acquisition, feature extraction and inference in a tiny embedded package. This plug-and-play mechanism eliminates the need for offline high-end expensive setups.   
\item \textbf{NN training on mobile phones:} We have deployed an on-device training methodology for mobile phones easing the work of the maintenance personnel in industries.
\item \textbf{Simultaneous data acquisition and inference:} 
Implemented concurrent data acquisition and inference to prevent loss of data samples.This is crucial for real-time health prognosis. This is achieved with the aid of DMA. 
\item \textbf{Generic architecture:} The infrastructure and algorithm can be extended to distinct types of actuators exhibiting similar types of transient response, thereby increasing the scope of applications.
\end{itemize}

\section{PreMa System Design Methodology} \label{sec:PreMa System Desing}

\begin{figure}[h]
  \centering
  \includegraphics[width=\linewidth]{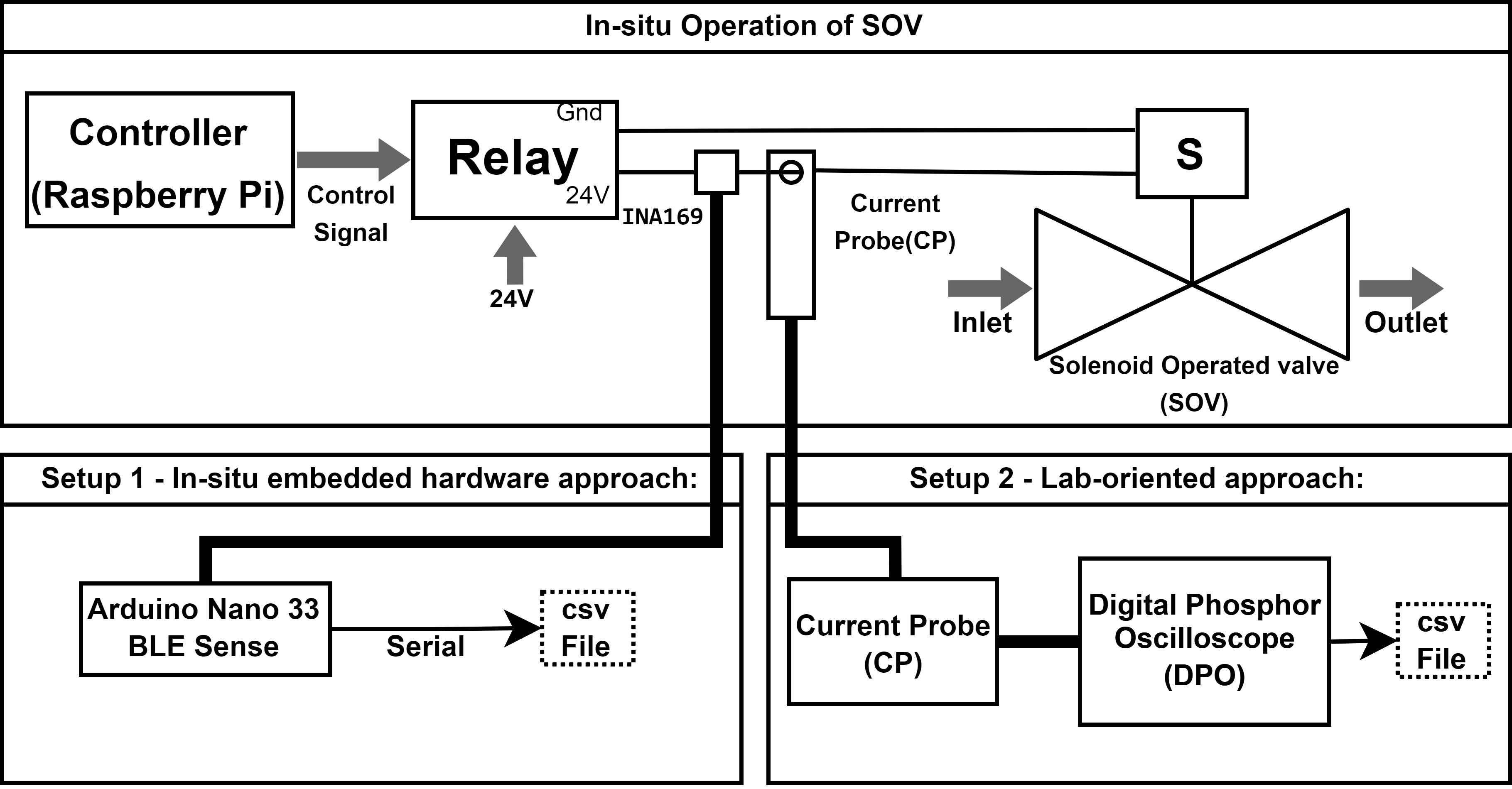}
  \caption{Data Acquisition Setup. }
  \label{fig:DataAcq}
  \Description{}
\end{figure}
\subsection{Data Acquisition}
Figure \ref{fig:DataAcq} elucidates two sensing setups, an in-situ measurement approach using embedded hardware (a.k.a Setup 1) and standard laboratory-oriented measurement approach (a.k.a Setup 2) for data acquisition. Data acquired is a time series of SV drive current. The region of interest (ROI) from the acquired SV drive current is SV's transient response during its excitation phase as it encapsulates all the electro-mechanical action occurring at its triggering phase.

\subsubsection{\textbf{Setup 1 - In-situ Embedded Hardware Approach:}}
\label{sec:setup 1}
In setup 1, we present a real-time, cost-effective, autonomous data acquisition system without compromising the quality of data which is expected to be at par with offline laboratory-oriented measurement approach. It utilises an Arduino Nano 33 BLE Sense (AN33BS) \cite{[25]}, which consists of low power nrf52840 system on a chip (SOC) . In order to measure the current with precision, high shunt current sensor INA169 \cite{[26]} is inculpated in the data acquisition circuit. The ROI lies within the 100ms (Frequency of ROI: 10Hz) of the acquired drive current. Hence, to attain faithful reconstruction 1KHz sampling frequency was chosen which is 50 times the Nyquist sampling frequency.

\begin{figure}[h]
  \centering
  \includegraphics[width=\linewidth]{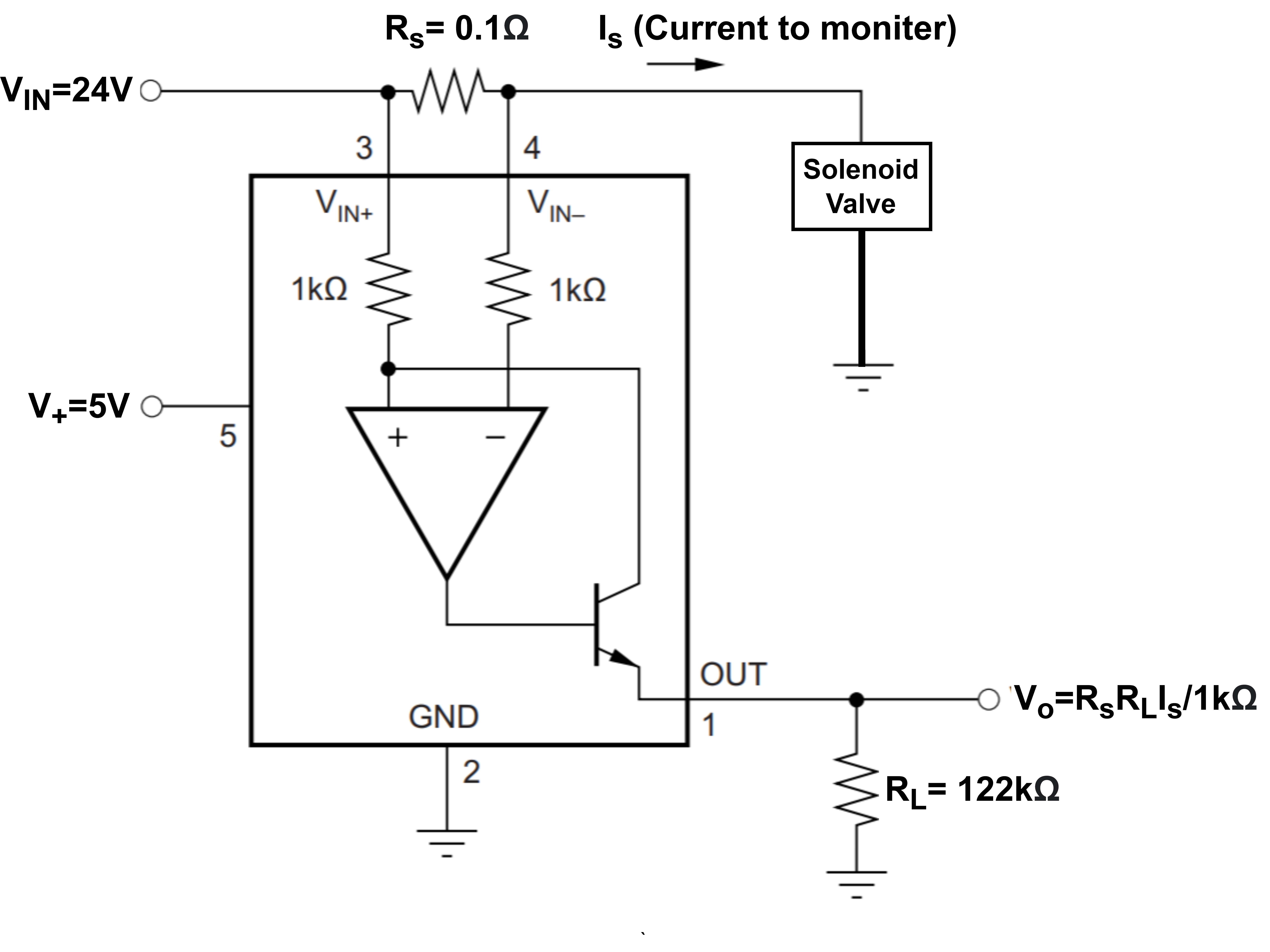}
  \caption{Design for SV's Current Senor. }
  \label{fig:INA169}
  \Description{}
\end{figure}

Figure \ref{fig:INA169} illustrates the design of current sensor to meet the operating condition of the SV. 
For 24V supply voltage applied to SV, maximum settling current was observed to be 250mA, with 20mA tolerance. The voltage obtained at the output end of INA169 `$V_O$' is calculated using equation \ref{eq:1}. The gain `$G$' and resistor values `$R_S$',`$R_L$' are chosen using equation \ref{eq:2}.

\begin{equation} \label{eq:1}
V_O = \frac{R_S.R_L.I_S}{1k\Omega}
\end{equation}

\begin{equation} \label{eq:2}
  G = \frac{R_S.R_L}{1k\Omega}
\end{equation}

To utilize the full scale reading (FSR) range of 12-bit Analog to Digital Converter(ADC) of AN33BS\\
$V_{O_{max}} = 3.3 V$ 

\begin{displaymath}
G = \frac{V_O}{I_S} = \frac{3.3V}{270mA} = 12.22
\end{displaymath} 
Hence,
\begin{displaymath}
  \frac{R_S.R_L}{1k\Omega}=12.22
\end{displaymath}

$R_S = 0.1\Omega, R_L= 122k\Omega$ was chosen.\\

Hence, the current sensor (INA169) with the required resistor combination is interfaced with the $12$ bit ADC of AN33BS to form an inexpensive yet reliable method of current measurement and data acquisition.

\subsubsection{\textbf{Setup 2 -Standard Laboratory-Oriented Approach:}}
\label{sec:setup 2}
\begin{figure}[h]
  \centering
  \includegraphics[width=\linewidth]{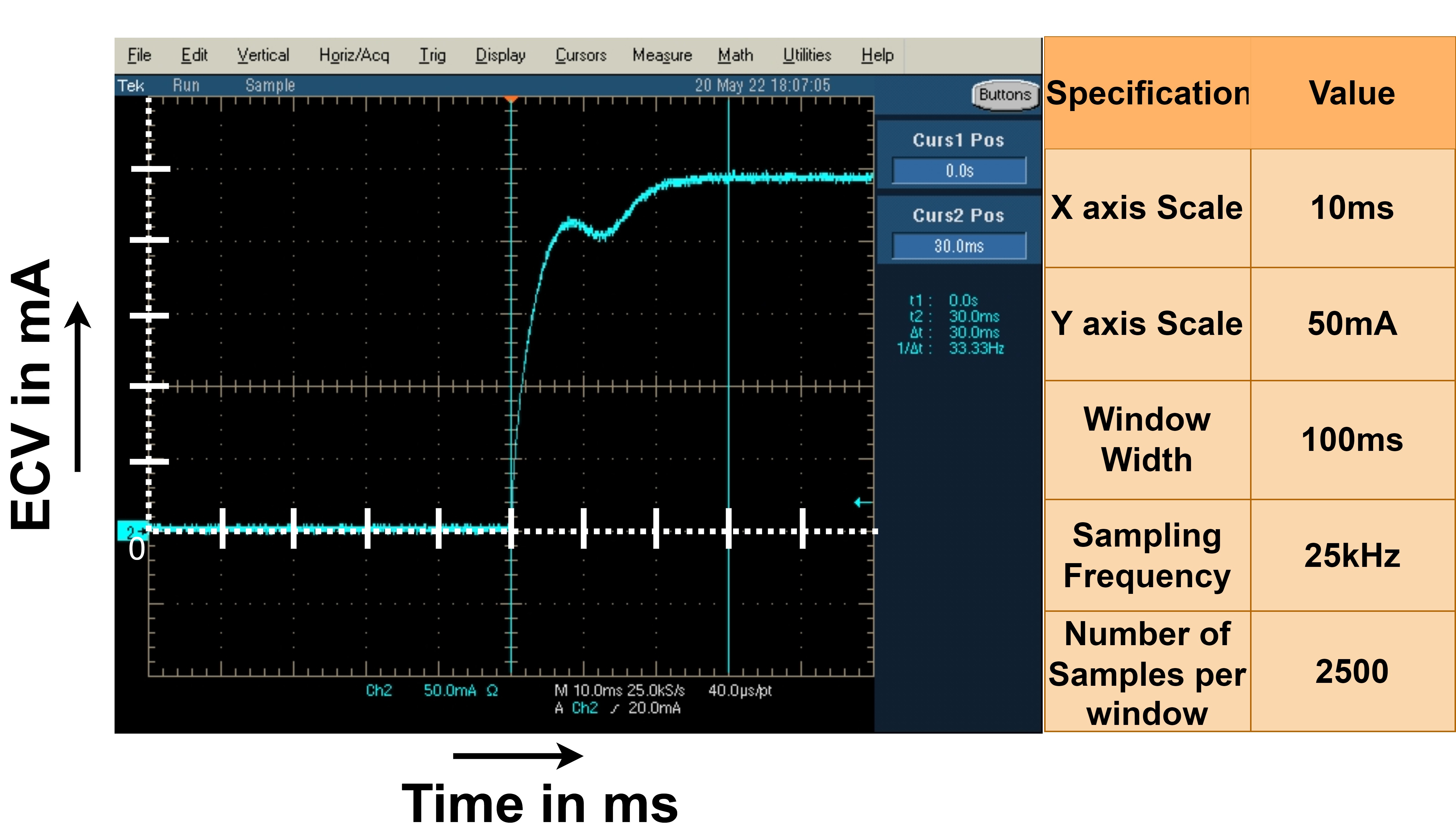}
  \caption{Transient Response Captured Via CP+DPO. }
  \label{fig:dacm2}
  \Description{}
\end{figure}

In this method, we employ several expensive laboratory instruments for acquiring transient current response of SV. We utilize a current probe (CP) Tektronix TCP$312$A, a current probe amplifier (CPA) Tektronix TCPA300 \cite{[23]} and a digital phosphor oscilloscope (DPO) Tektronix TDS5104 \cite{[24]}. The CP is clamped to the $24$V supply line of the SV and by tuning the settings of DPO to our required specification, the transient response of SV during its excitation phase was precisely extracted in the frame. Figure \ref{fig:dacm2} projects the variation of SV's electric current value (ECV) or drive current over time during the excitation phase of SV, captured on the DPO System.

This setup captures the 100ms ROI of the current profile automatically whereas in setup 1 this ROI is obtained by applying an rising edge algorithm which is explained in section \ref{sec:FeatureExtraction}. The acquired ROI is then converted to CSV format for further processing.

\subsection{Fault Analysis and Need for Autonomous, Cognitive, Standalone, Real-Time Embedded System}


\begin{figure}[h]
  \centering
  \subfloat[Good Valve.]{\includegraphics[width=0.5\linewidth]{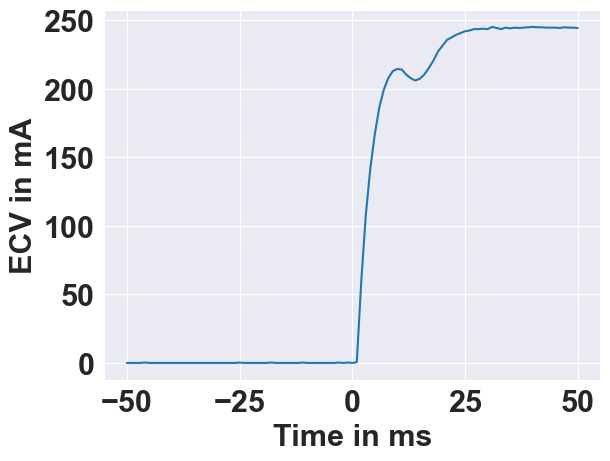}\label{fig:f1}}
  \hfill
  \subfloat[Spool Stuck.]{\includegraphics[width=0.5\linewidth]{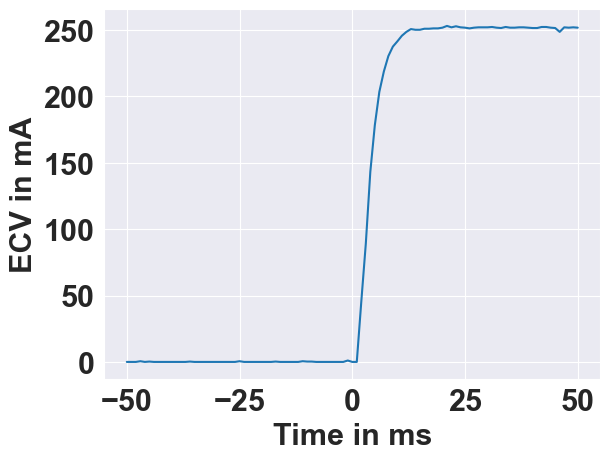}\label{fig:f2}}
  
    \subfloat[Spring Failure.]{\includegraphics[width=0.5\linewidth]{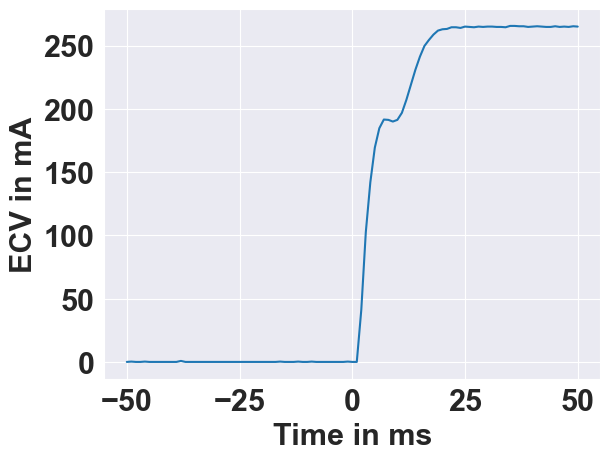}\label{fig:f3}}
  \hfill
  \subfloat[Under Voltage.]{\includegraphics[width=0.48\linewidth]{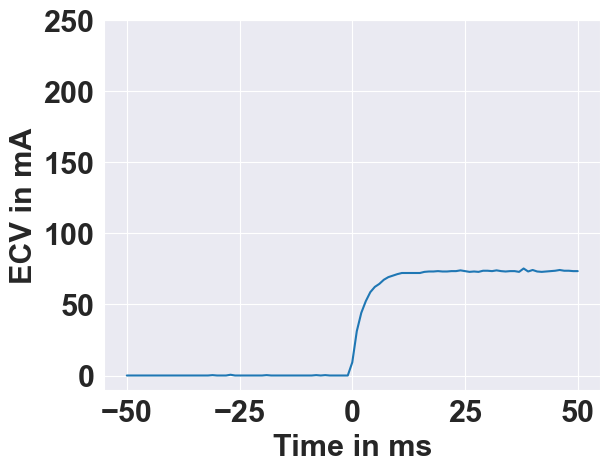}\label{fig:f4}}
  \caption{SV Faults and Respective Transient Current Response.}
  \label{fig:faults}
\end{figure}

As discussed in section \ref{section:intro}, the two factors contributing towards the failure of a SV are the internal components degradation and external factors. Our measurement campaign for fault detection involves the computation of transient response of the SV under four different conditions. Figure \ref{fig:faults} illustrates the four different transient responses obtained via setup \ref{sec:setup 1}. The transient responses are of (a) A good valve, (b) Spool stuck valve, (c) Valve with a spring failure and (d) A valve experiencing under voltage. This transient response of all the states is captured during the actuation phase of the valve. The current profile thus extracted encompasses all the characteristics the SV exhibits during its transition from idle to excitation state. Hence, capturing transient response for different types of faults lays a firm foundation for analyzing the SV. It is evident from the graph that considering the transient response for feature extraction to facilitate the failure analysis is considered a well-rounded choice as it shows clear distinction between the various valve failures.

In order to extract the transient response from the SV drive current, we have developed an embedded system design for data acquisition, which eliminates the need of high-end laboratory equipment. Inference of the valve health and failure classification needs an autonomous system which can be facilitated by a deep learning framework. The attributes extracted from the data acquisition module can be fed into a deep learning pipeline in an offline mode. However, this approach lacks real-time inference. As a fortification to this drawback, we propose a real-time, cognitive, in-situ, standalone embedded package. This architecture facilitates real-time data acquisition, feature extraction followed by inference by NN models at the edge level. This system also has the capability of training the NN on mobile phones.

\subsection{Feature Extraction} \label{sec:FeatureExtraction}
\begin{figure}[h]
  \centering
  \includegraphics[width=\linewidth]{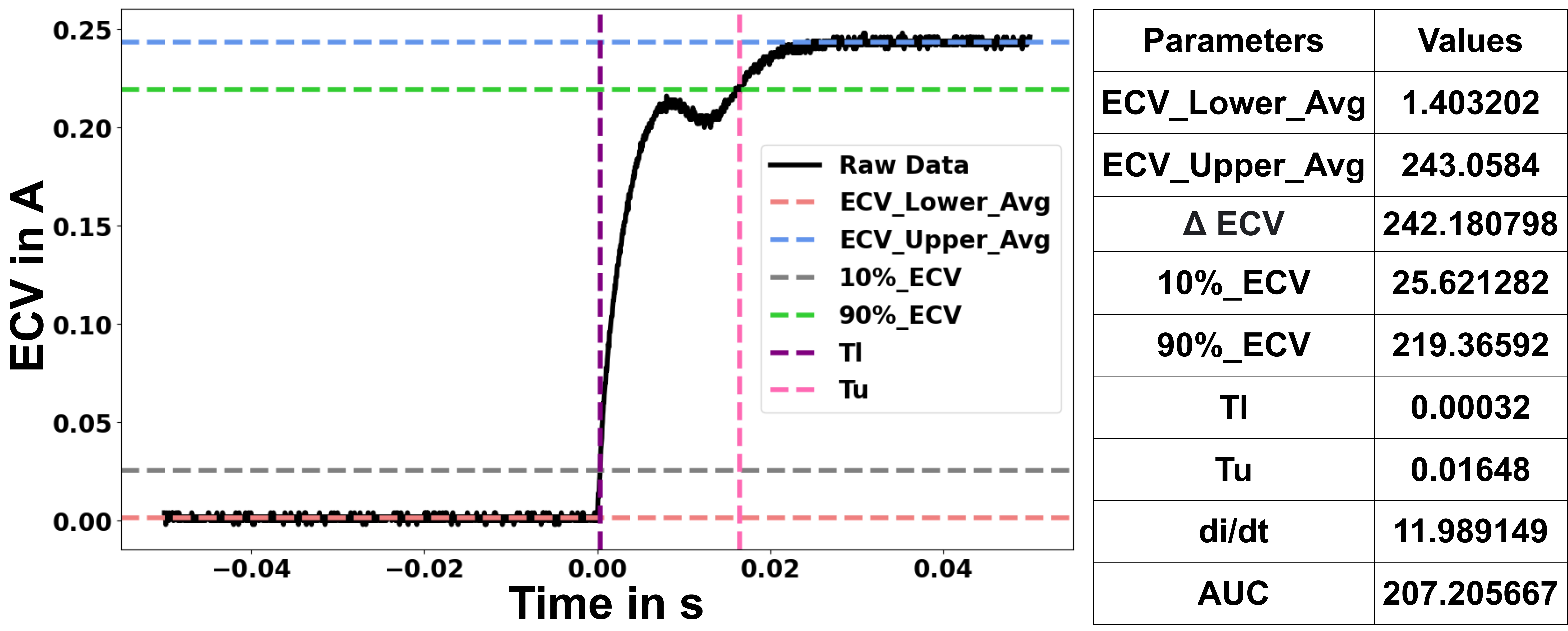}
  \caption{Extracted Features of SV's Transient Response}
  \label{fig:FE}
  \Description{}
\end{figure}

A SV's current signature conveys crucial information about its present state of health. It has been proved that current transient response during this phase exhibits a distinct profile with distinct faults. The raw data which we fetched is in the form of ECV of SV with respect to time during the actuation phase of SV. Extracting the solenoid drive current is not sufficient for training purposes. Henceforth, we need to draw out the key attributes from the current profile during its triggering phase. Therefore, feature engineering is employed for transforming raw data into meaningful features that offer a better shape for training. In this section, we present a unique way of feature extraction which yields features similar to the attributes from the step response of a dynamic model system \cite{[37]}. Algorithm \ref{alg:1} explains the procedure for feature extraction. The features which we extracted are ECV lower average (ECV\_Lower\_Avg) which is the average of ECVs for a duration of 50ms before actuation, ECV upper average (ECV\_Upper\_Avg) is the average of ECVs for a period of 20ms after actuation, $\Delta ECV$ is the difference between ECV\_Upper\_Avg and ECV\_Lower\_Avg, 10\%\_ECV is the summation of $\Delta ECV$ and 10\% of ECV\_Lower\_Avg , whereas 90\%\_ECV is the summation of $\Delta ECV$ and 90\% of ECV\_Lower\_Avg, Tl is the time at which ECV of SV reaches 10\%\_ECV, whereas Tu represents time when ECV od SV achieves 90\%\_ECV, slope (di/dt ) is the rate of rise in current from Tl to Tu and AUC is the weighted area of the transient response during the actuation phase. Figure \ref{fig:FE} visually depicts the important features that are extracted from the transient response. In the figure \ref{fig:FE}, black line represents the raw data, the red line represents the ECV\_Lower\_Avg, the blue line represents the ECV\_Upper\_Avg and the grey line represents 10\%\_ECV. The green line, the purple line and the pink line represents 90\%\_ECV, Tl and Tu respectively.

\RestyleAlgo{ruled}
\SetKwComment{Comment}{/* }{ */}

\begin{algorithm}[hbt!]
\small
\caption{Feature Extraction for Setup 2}\label{alg:1}
\textbf{Input:} buffer\_data\\
$i \gets 5$\;
$count \gets 0$\;
\While{$i \le (buffer\_size-1)$}{
    $window\_average \gets Mean(buffer\_data[i-5:i])$\;
    {\If{$window\_average \geq 40$ and $buffer\_data[i-5] \leq 5$}{
      $zero\_index \gets i-5$\;
      $ECV\_Lower\_Avg \gets Mean(buffer\_data[zero\_index-50:zero\_index])$\;
      $ECV\_Upper\_Avg \gets Mean(buffer\_data[zero\_index+30:zero\_index+50])$\;
      $\Delta ECV \gets ECV\_Upper\_Avg - ECV\_Lower\_Avg$\;
      $10\%\_ECV \gets \Delta ECV * 0.1 + ECV\_Lower\_Avg$\;
      $90\%\_ECV \gets \Delta ECV * 0.9 + ECV\_Lower\_Avg$\;
      {\For{$j \gets zero\_index-50$ to $min(zero\_index+100,buffer\_size-1)$}{
        {\If{$buffer\_data[j] \geq 10\%\_ECV$}{
          $Tl \gets count-50$\;
          break\;
        }
        $count \gets count + 1$\;
        }
      }
      }
      {\For{$k \gets min(zero\_index+100, buffer\_size-1)$ to $j$}{
        {\If{$buffer\_data[k] \leq 90\%\_ECV$}{
          $Tu \gets 100 - count + 1$\;
          break\;
        }
        $count \gets count + 1$\;
        }
      }
      }
      $di/dt \gets (90\%\_ECV - 10\%\_ECV)/(Tu - Tl)$\;
      $AUC \gets ((buffer\_data[zero\_index] + buffer\_data[zero\_index+ 30])/2 + sum(buffer\_data[zero\_index+1:zero\_index+30]))/30$\;
      $i \gets i+30$\;
    }
  }
  $i \gets i+1$
}
\end{algorithm}

		
            
            
            
            
            
            
            
            
            

One significant step is to locate the most appropriate window of current profile. For capturing this window, the position of rising edge of the profile is required to be detected. For this purpose, we employed a moving average algorithm with window size of 5ms worth of samples. When the average of the window reaches a threshold of 40mA (around 15\% of maximum settling current) and the starting value of window is less than 5mA, a rising edge is located. Then we centered a frame of 100ms around this point for computing the above-mentioned attributes. Our feature extraction algorithm is deployed on the embedded platform for real-time feature extraction. The main takeaway from our algorithm is that di/dt inherits all of the above-mentioned features characteristics, thereby acting as a binding capsule. Additionally, AUC captures crucial trends in SV's transient response which is useful in RUL estimation.

\subsection{Comparison Between the Two Setups }

\begin{figure*}[t]
  \centering
  \captionsetup{justification=centering}
  \subfloat[Comparison of reconstructed transient response of the two setups]{\includegraphics[width=0.33\linewidth]{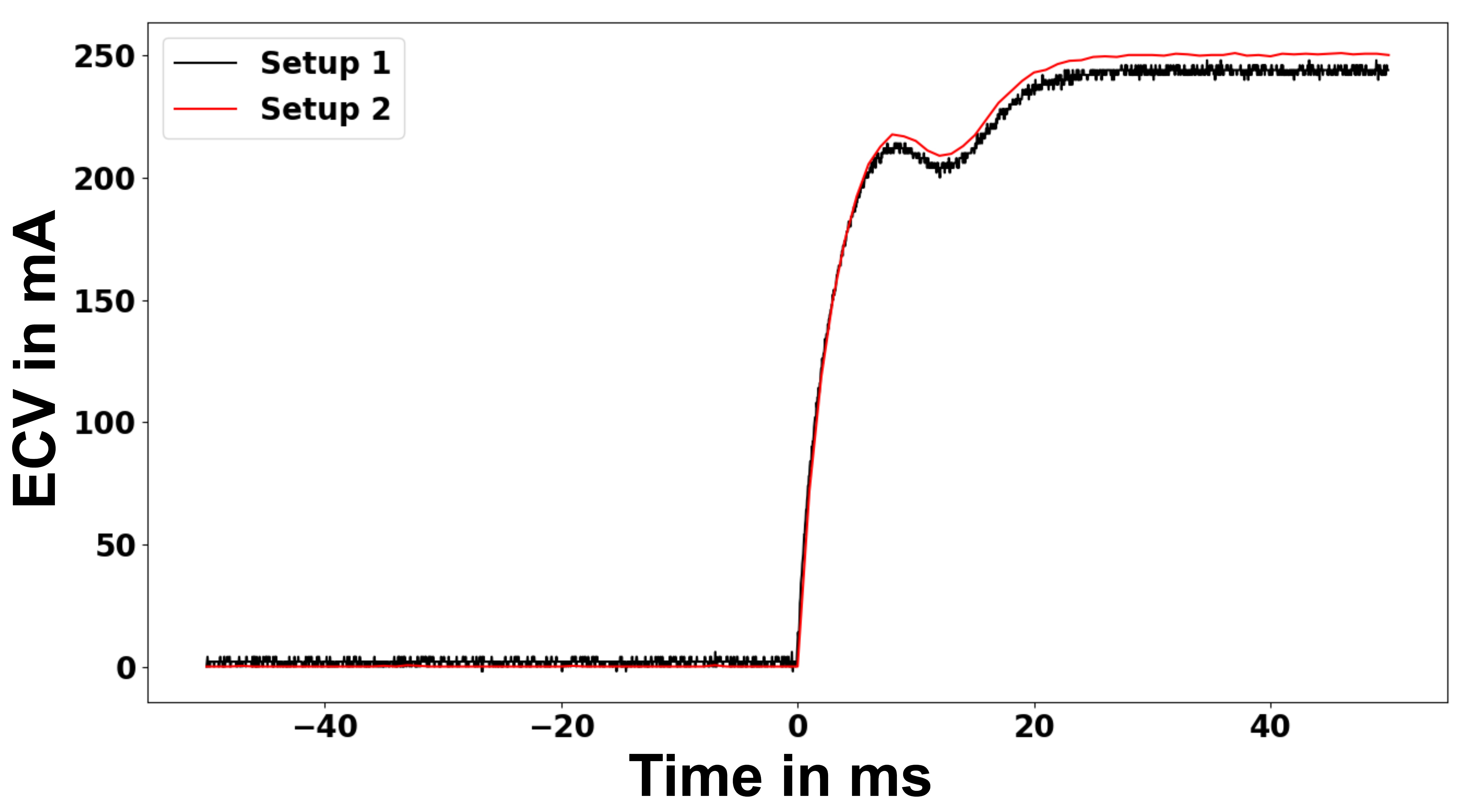}\label{fig:s1_vs_s2}
}
\hfill
  \subfloat[Variation of di/dt of both setups\label{fig:5b}]{\includegraphics[width=0.33\linewidth]{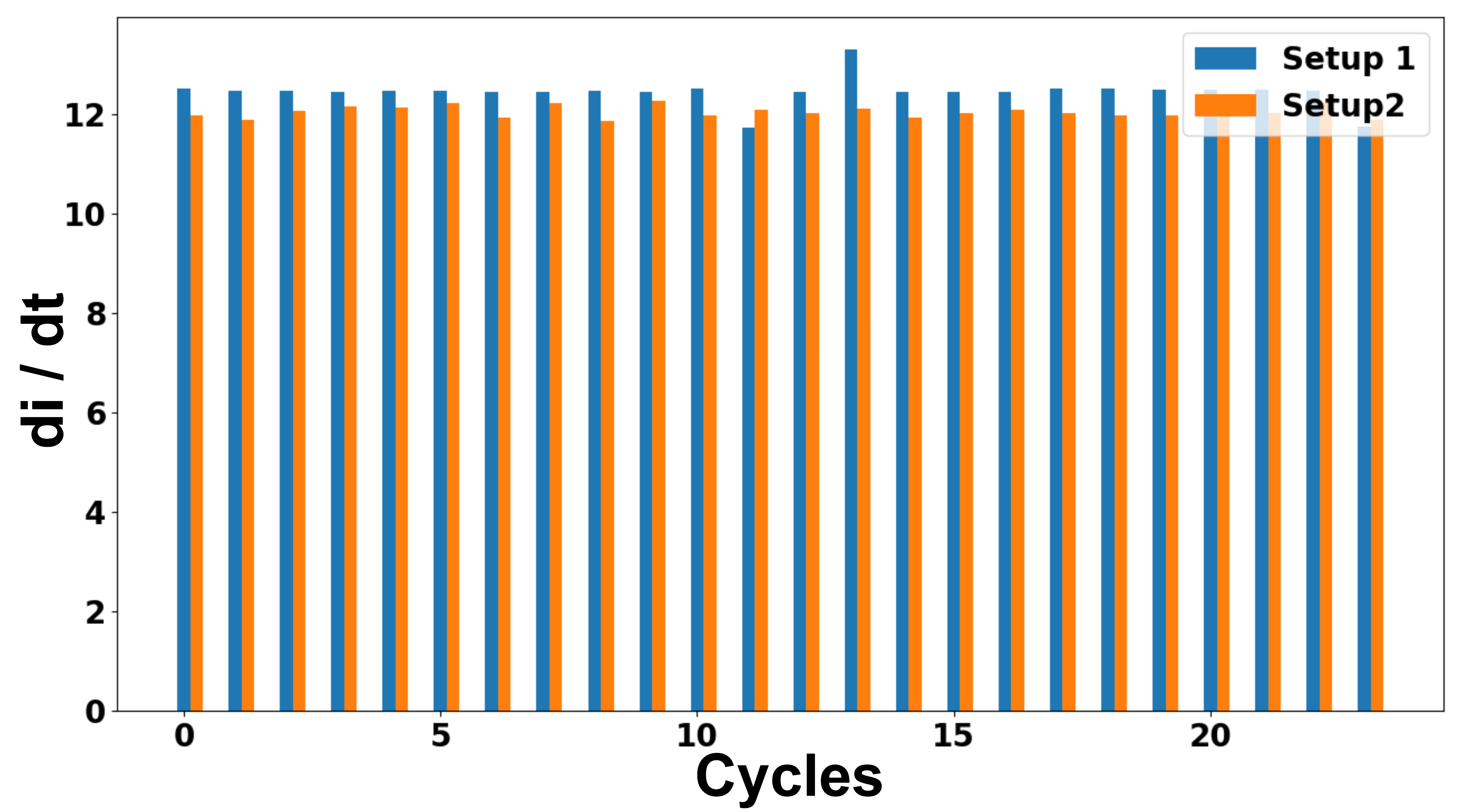}}
 \hfill
  \subfloat[Variation of AUC of both setups\label{fig:5c}]{\includegraphics[width=0.33\linewidth]{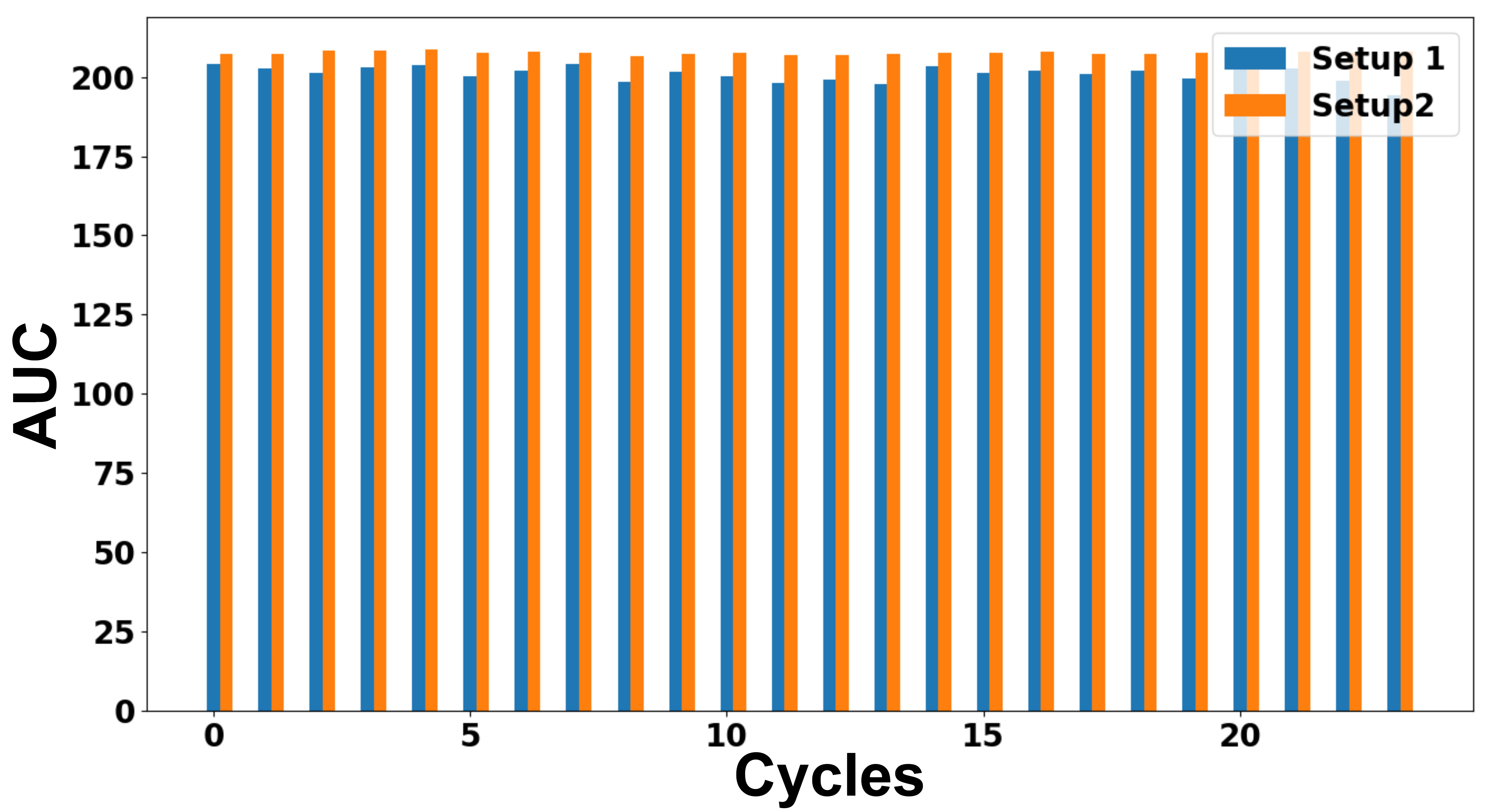}}
  \caption{Comparison of Both Setups}
\end{figure*}

\begin{table}[h]
\small
  \caption{Comparison Between the Two Setups}
  \label{tab:comparison}
  \begin{tabular}{ c c c }
    \toprule
    Parameter&Setup 1&Setup 2\\
    \midrule
    Cost& Low Cost & Very Expensive\\
    Security & More Secure & Less Secure$,$\\ 
    & data stays local & data needs to be\\
    &  & transferred\\
    Ease Of Usability & Easy$,$in$-$situ & Difficult for in$-$situ\\
    &  & measurement\\
    Scalability & Easily Scalable & Less Scalable\\
    Ability To Automate & Easy To Automate & Manual Process\\
  \bottomrule
\end{tabular}
\end{table}

Table \ref{tab:comparison} compares the two setups based on a number of factors, emphasising the importance of our design. The standard approach as discussed in setup 2 (Section \ref{sec:setup 2}) sets high quality standards in terms of current probe's sensing accuracy, sampling rate, adaptable settings, and storage capabilities. However, the process is extremely tedious, expensive and it is onerous to capture the in-situ transient response of every excitation in real-time. In contrast, setup 1 (Section \ref{sec:setup 1}) balances the trade off between cost and precision of the system.


  



Figure \ref{fig:s1_vs_s2} presents the reconstructed graphs for the same transient response of the valve captured via setup 1 and setup 2. Figure \ref{fig:5b} and \ref{fig:5c} depicts the variation in the features extracted from 24 transient responses captured via setup 1 and setup 2 . Table \ref{tab:Accuracy Comprasion} clearly shows that results obtained from setup 1 are in-line with setup 2 with minor variation. From table \ref{tab:Accuracy Comprasion}, it can be substantiated that both the setups yield almost equivalent attributes which is evident from the mean absolute error (MAE) and root mean square error (RMSE) between the features extracted through the two setups. Hence, we are not only able to mimic the standard laboratory-oriented setup reduced to an in-situ embedded hardware setup but also able to improve on various aspects mentioned in the Table \ref{tab:comparison} by significant proportions, hence making setup 2 redundant.


\begin{table}
\small
  \caption{Accuracy Comparison Between the two Setups}
  \label{tab:Accuracy Comprasion}
  \begin{tabular}{p{0.2\linewidth}  p{0.3\linewidth} p{0.3\linewidth}}
    \toprule
    Features & Mean absolute error (MAE) & Root Mean Square Error (RMSE)\\
    \midrule
    di$/$dt & 0.455787 & 0.492007\\
    AUC & 6.885348 & 7.108937\\ 
  \bottomrule
\end{tabular}
\end{table}

\subsection{NN Model Construction for Fault Detection}
\subsubsection{\textbf{Experimental Setup and Data set description}}\label{sec:FDExperimentalSetupAndDataset}


\begin{figure}[t]
  \centering
   \captionsetup{justification=centering}
  \subfloat[Fault Detection Experimental Setup]{\includegraphics[width=0.535\linewidth]{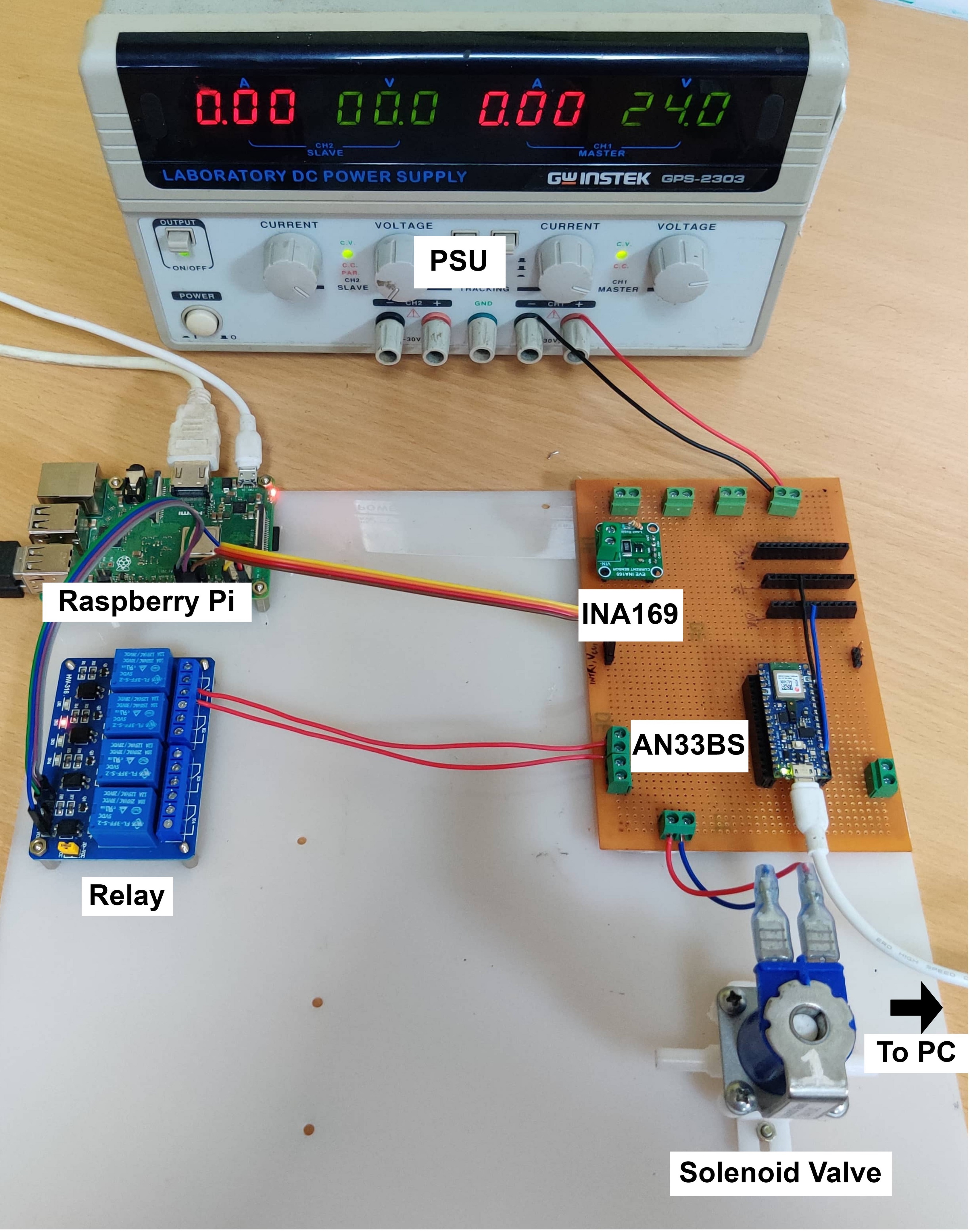}\label{fig:FDataAcq}}
  \hfill
  \subfloat[RUL Experimental Setup]{\includegraphics[width=0.437\linewidth]{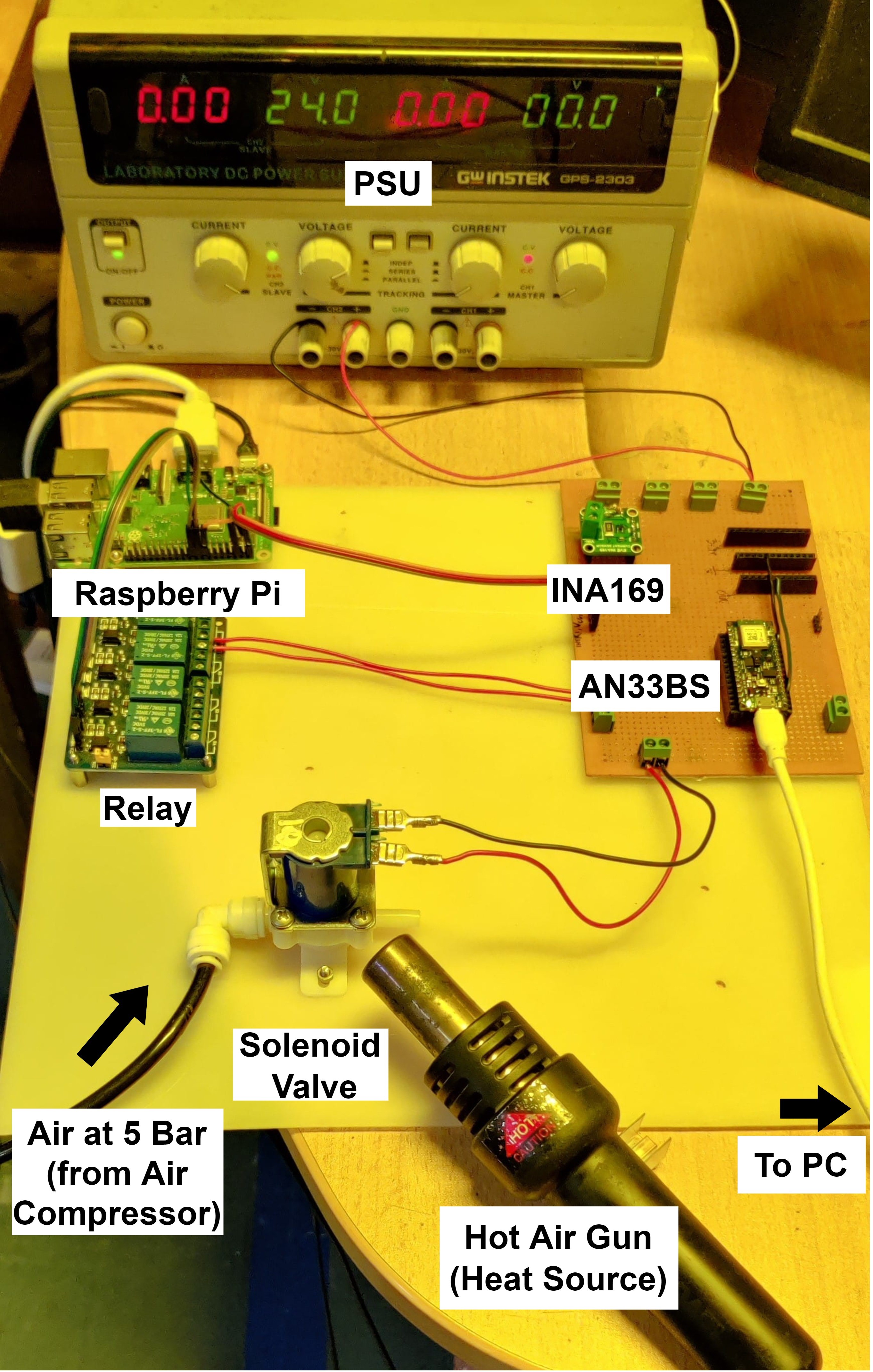}\label{fig:RULSetup}}\\
\caption{Fault Detection and RUL Experimental Setup}
\end{figure}

 Figure \ref{fig:FDataAcq} shows the setup that consists of our in-situ embedded hardware approach (Setup 1) for SV's data acquisition. The experiment was conducted on 9 different valves from different manufacturers to diversify the data set. It consisted of 6 SVs in good condition, one with spool stuck, one with spring failure and one good SV which was operated with an under voltage range of 8V-14V. All SVs except under voltage condition were operated for a total of 1000 cycles, acquiring the time series transient response for every 5th operation. For under voltage condition, the SV was operated for 500 cycles for every 2V increment, acquiring the transient response for every 5th operation. Throughout the experiment, the SV was operated at a frequency of 0.5Hz. The dataset is composed of AUC and di$/$dt as discussed in section \ref{sec:FeatureExtraction}. The data set totally comprised of 600 samples for good valve, 200 samples for spool stuck, 200 samples for spring failure and 400 number of samples for under voltage conditions.

\subsubsection{\textbf{Constructing the NN Model}}

The developed fault detection NN has a feed forward architectural style. Our NN is made up of a two-dimensional input layer with 36 neurons for the extracted characteristics (di/dt and AUC), two hidden layers with 24 and 12 neurons. The output layer has 4 neurons, which represents the number of categories (3 faults and 1 no fault) we wish to detect. The input and hidden layers employ the Leaky ReLU activation function, whereas the output layer uses softmax. Table \ref{tab:fd_archi} represents the architecture of the model. The optimizer we used is RMSProp and the loss function we employed is categorical cross entropy represented in the equation \ref{eq:3}, where $y_i$ is the actual value and $\hat{y}_i$ is the predicted output.

\begin{equation}\label{eq:3}
    LOSS=\sum_{i=1}^{n} y_i\cdot log(\hat{y}_i)
\end{equation}


\begin{table}
\small
  \caption{Fault Detection Model Architecture}
  \label{tab:fd_archi}
  \begin{tabular}{p{0.33\linewidth}  p{0.2\linewidth} p{0.2\linewidth}}
  \multicolumn{3}{c}{Model: "Sequential"} \\
    \toprule
    Layer (Type) & Output Shape & Parameters\\
    \midrule
    dense (Dense) & (None, 36) & 108\\
    dense\_1 (Dense) & (None, 24) & 888\\
    dense\_2 (Dense) & (None, 12) & 300\\
    dense\_3 (Dense) & (None, 4) & 52\\
  \bottomrule
\end{tabular}
\end{table}

\subsection{NN Model Construction for RUL Estimation}\label{sec:NNRUL}

\subsubsection{\textbf{Experimental Setup and Data set description},\label{sec:RULExperementalSetup}}

 ---Figure \ref{fig:RULSetup} depicts the run-to-failure setup which consists of the SV, a heating source, our data acquisition system and an air compressor. Since, the SV is rated up to 8 bar, the compressor provides the required pressure at the inlet of the SV. For the run-to-failure experiment, the device under test (DUT) was subjected to the following conditions: (a) Switching operations at 0.5 Hz, (b) 60\% of rated pressure (5Bar) and (c) Temperature higher than ambient temperature using the heat source. The actuation current profile was captured for every 5 SV operation cycles. At room temperature(26°C), the SV was operated for a duration of 1000 cycles and 100 cycles for every subsequent increment of valve temperature by 10°C. The features (di/dt and AUC) extracted from the actuation current response during the aforementioned conditions serves as the dataset for the NN model.


\subsubsection{\textbf{Constructing NN Model}}
Table \ref{tab:rul_archi} projects the NN regression model architecture employed for estimating RUL. The dense input layer of the model has a size of two, and the activation function for this layer was ReLU. There are 64 neurons in the input layer. There are two hidden layers in the model, one with 16 neurons and the other with four. The hidden layers used the ReLU activation function. The regression output was represented by a single neuron in the output layer. The model utilizes RMSProp optimizer and loss function is MAE shown in equation \ref{eq:4}, where $y_i$ is the actual value, $\hat{y}_i$ is the predicted output and $n$ is the total number of data samples. 

\begin{equation} \label{eq:4}
LOSS=\frac{\sum_{i=1}^{n}|y_i-\hat{y}_i|}{n}
\end{equation}

\begin{table}
\small
  \caption{RUL Estimation Model Architecture}
  \label{tab:rul_archi}
  \begin{tabular}{p{0.33\linewidth}  p{0.2\linewidth} p{0.2\linewidth}}
  \multicolumn{3}{c}{Model: "Sequential"} \\
    \toprule
    Layer (Type) & Output Shape & Parameters\\
    \midrule
    dense (Dense) & (None, 64) & 192\\
    dense\_1 (Dense) & (None, 16) & 1040\\
    dense\_2 (Dense) & (None, 4) & 68\\
    dense\_3 (Dense) & (None, 1) & 5\\
  \bottomrule
\end{tabular}
\end{table}

\subsection{Converting Models for On-Device Training and Inference}
Once neural architecture system design has been developed, it needs to be converted to proper format for it to be utilized for on-device training and inference. We utilize the TFLite open source package for proper formatting of the developed NN architecture.


TensorFlow Lite \cite{[38]} is a collection of tools that lets programmers deploy their machine learning models on embedded, mobile, and other edge platforms, enabling on-device machine learning. In addition to executing inference, TensorFlow Lite also allows you to train your models on mobile phones \cite{[39]}.
On-device training overcomes five important barriers for our application: (a) latency (no need to transfer data to a server and then wait for responses), (b) privacy (no confidential information leaves the system), (c) connectedness (no network link is required), (d) storage (smaller model ), and (e) low energy usage.

\begin{figure}[h]
  \centering
  \includegraphics[width=0.8\linewidth]{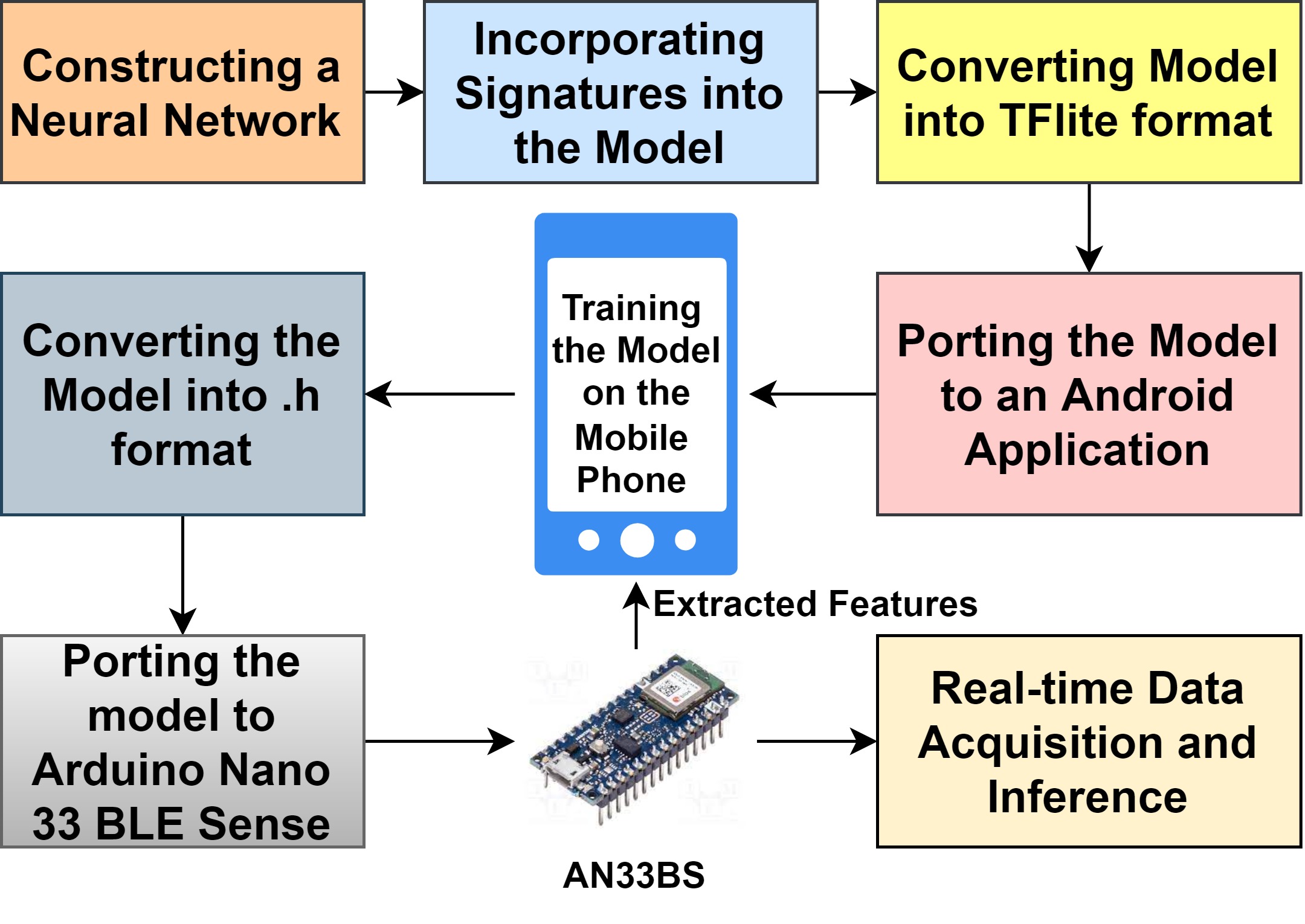}
  \caption{On-Device Training and Inference Flow }
  \label{fig:OnDevice}
  \Description{}
\end{figure}

In this work, we deploy an end-to-end framework for real-time predictive maintenance. We have implemented a novel way of on-device training on mobile phones and on-device inference on AN33BS. This entire approach eliminates the need of offline training and inferencing on powerful and expensive PCs. Figure \ref{fig:OnDevice} illustrates the consolidated steps to implement the above-mentioned approach.

We employed four functions to our existing TensorFlow NN model: (a) train, (b) infer, (c) save, and (d) restore. These functions will be uncovered to signatures when it is converted to tflite model format by TensorFlow Lite converter (an optimized Flat Buffer format identified by the .tflite file extension). For instance, the train function performs the back propagation process using automatic differentiation operations, and updates model parameters. These kinds of signatures can be interpreted in real-time by tflite interpreter instance on android platforms to perform on-device training. We acquire the features of transient response of SV from the AN33BS in real-time for training the model on mobile phone and send back the updated model in library package to the embedded platform enabling it to perform on-device inference. In regards to on-device inference on AN33BS, tlfite model obtained must again be converted to c byte array loaded into header file. This can be invoked by tflite interpreter in real-time aiding on-device inference\cite{[40]}.

\subsection{Concurrent real-time Data Acquisition and Inference}

\begin{figure}[h]
  \centering
  \includegraphics[width=0.85\linewidth]{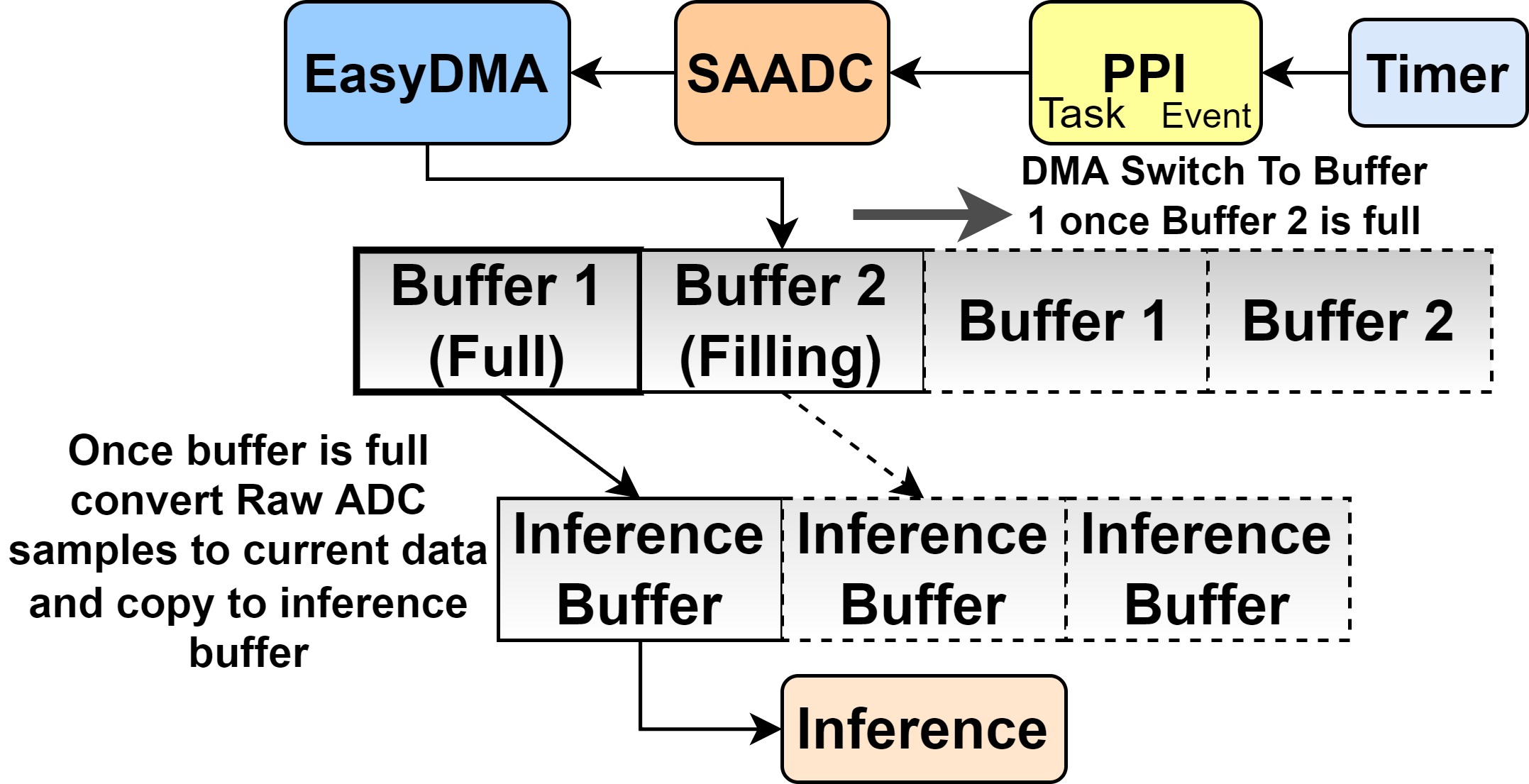}
  \caption{Ping Pong Buffer Process Flow }
  \label{fig:PPBProcessFlow}
  \Description{}
\end{figure}

 \begin{figure*}[t]
  \centering
  \includegraphics[width=0.9\linewidth]{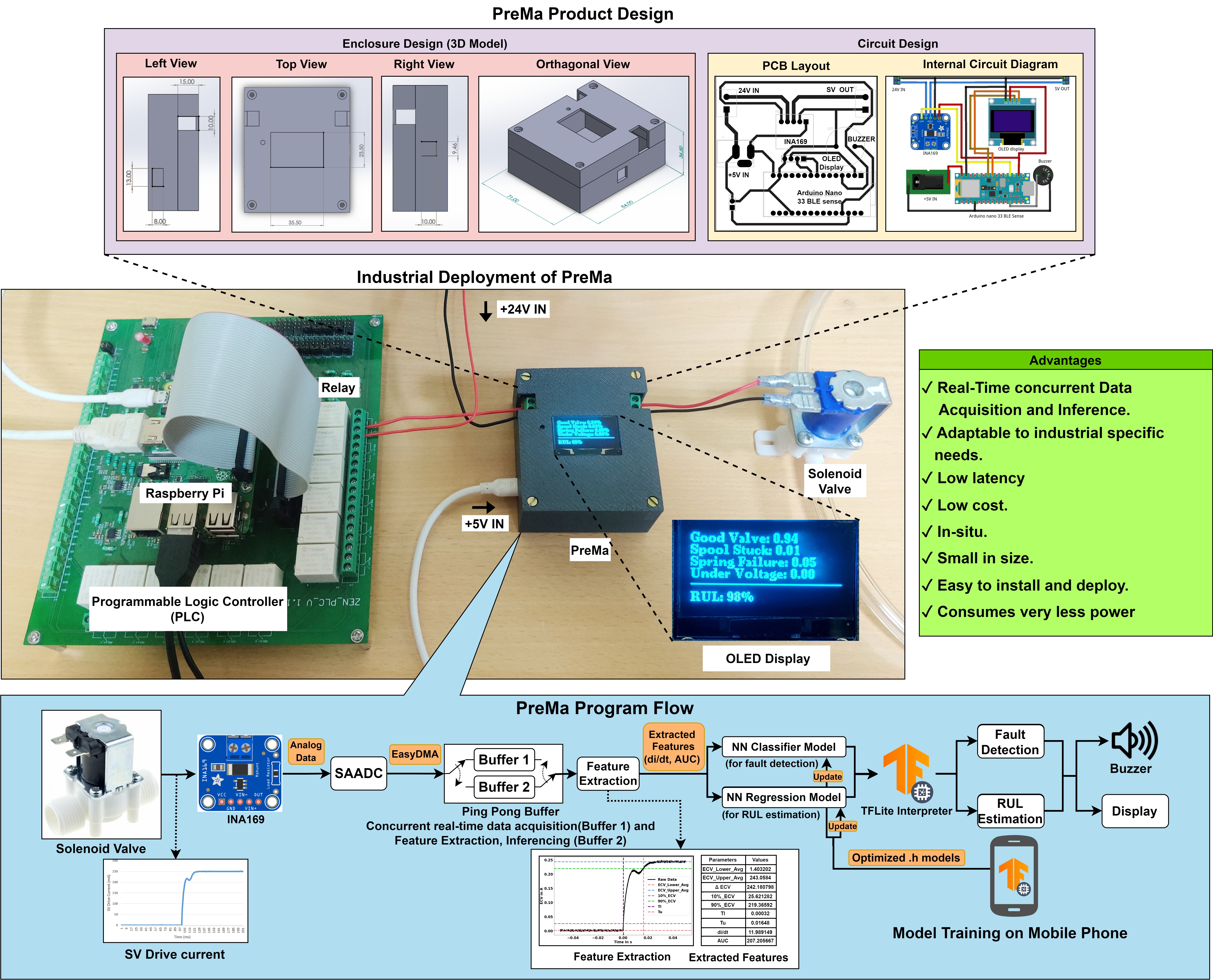}
  \caption{PreMa Product design, Industrial deployment and System Flow}
  \label{fig:PreMaDeployment}
  \Description{}
  \label{fig:PreMaDeployment}
\end{figure*}

Data acquisition and feature extraction followed by inference for predicting SV's state of health should proceed concurrently without loss of data during the inference stage. Figure \ref{fig:PPBProcessFlow} shows the deployment of ping-pong buffer \cite{[41]} with direct memory access (DMA) to facilitate concurrency. Nrf52840's (AN33BS) internal timer is connected to the successive approximation analog to digital converter (SAADC) via programmable peripheral interconnect (PPI) to enable us to sample data at the required sampling rate. EasyDMA is utilized to transfer the sampled analog data to buffer 1 of the ping pong buffer. When the buffer 1 is full it raises an Interrupt. The Interrupt Service Routine (ISR ) is responsible for switching buffer 1 to buffer 2 and raises a flag informing the processor to initiate the inferring process on the filled buffer 1. Once the flag is raised the processor converts the raw ADC samples into the current values and starts processing, while easyDMA is filling the sampled analog data into buffer 2 of the ping pong buffer, and the process repeats. Consequently, one buffer is filled while the other buffer is inferred thus, reaffirming the goal of parallel data acquisition and inference. 
 
\section{PreMa in real-time industrial deployment}
In this section, industrial deployment of PreMa is discussed. Figure \ref{fig:PreMaDeployment} projects the end-to-end life cycle of PreMa when deploying in production. The deployment process can be broadly classified in two stages namely (a) Product design \& packaging and (b) System flow. Product design is the primary step of product deployment. The product packaging consists of the application specific circuit design and 3D enclosure design. The application specific circuit design integrates the embedded hardware (AN33BS) with the current sensor (i.e INA169) and the alerting system consisting of an OLED display and a buzzer. This integration was then converted to a printed circuit board connecting all the aforementioned components which was then enclosed in the 3D printed enclosure. The system flow starting from sensing to inference was elaborated in section \ref{sec:PreMa System Desing}. In industrial deployment, the designed product and the system flow is integrated, which is our proposed product PreMa. PreMa is connected in series between the PLC and  SV  without interfering with the valve’s functionalities. In a real-time industrial environment, when a fault is detected or the predicted RUL drops below the threshold, an alarm is issued by triggering the buzzer. The product utilises the OLED screen to display the fault probabilities and the RUL continuously in real-time.

\section{Results, Discussions and Evaluation}

\subsection{Fault Detection of SV } \label{sec:FaultDetectionOfSV}

In this section, we present an in-depth examination of the end-to-end results pertaining to fault detection starting from on-device training on mobile phone to real-time on-device inference on AN33BS.

\subsubsection{\textbf{On-device training} \label{section:fd_train}}

\begin{figure}[h]
  \centering
  \includegraphics[width=\linewidth]{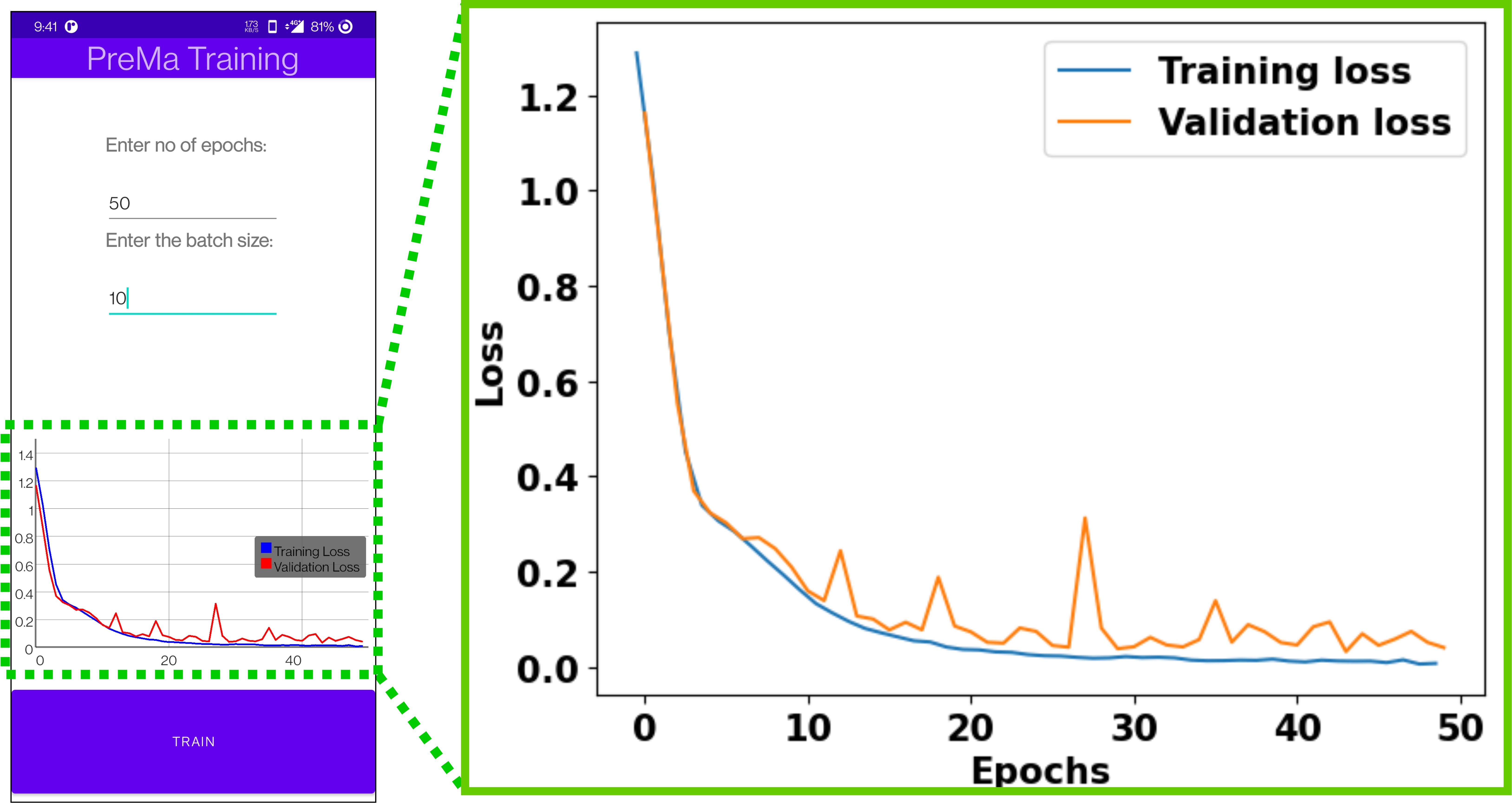}
  \caption{Fault Detection Training Loss Curve on Mobile Phone}
  \label{fig:FDLossCurve}
  \Description{}
\end{figure}

For training the NN model, we used data set presented in section \ref{sec:FDExperimentalSetupAndDataset}. The model is trained for 70\%, validated for 20\%, and, tested for 10\% of input dataset. Multiple combinations of hyper parameters i.e epochs(30, 50, 80, 100), batch size (5, 10, 30, 50) and optimizers (Adam, RMSProp, SGD) were experimented to find the best fit for the model. With the right combination of hyper parameters (epochs = 50, batch size = 10, optimizer = RMSProp), we are able to achieve 96.45\% accuracy on the test dataset. The model is trained on OnePlus 7 mobile phone \cite{[43]}. Figure \ref{fig:FDLossCurve} projects the on-device training and the trend of training and validation loss over the epochs.

\subsubsection{\textbf{On-device inference}}
\begin{figure}[h]
  \centering
  \includegraphics[width=0.85\linewidth]{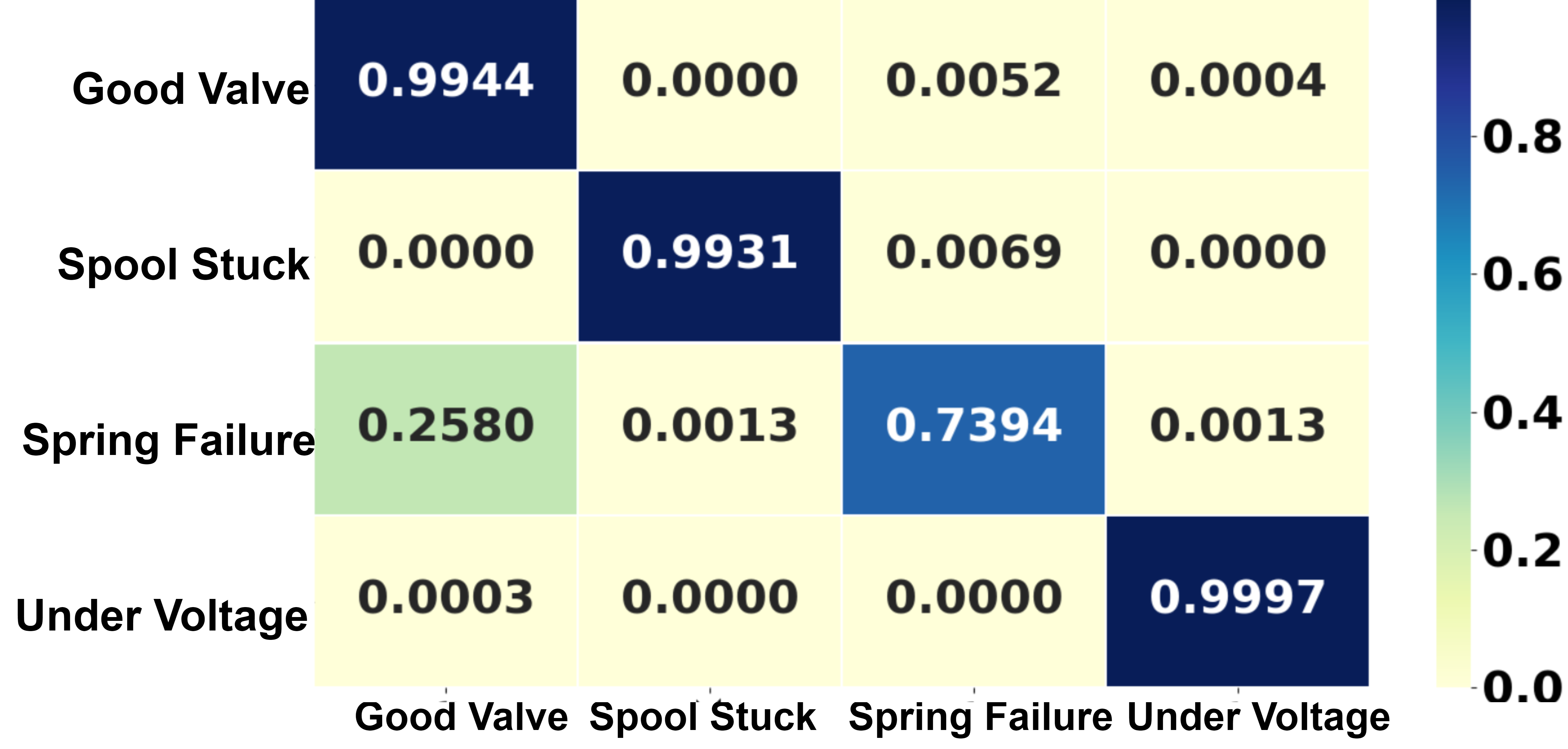}
  \caption{Classification Probability}
  \label{fig:HeatMap}
  \Description{}
\end{figure}

Figure \ref{fig:HeatMap} projects the performance of our model on production data of unseen SVs in the form of a heat map. For the test, 4 SVs with the 4 different conditions were analysed using PreMa. Each valve was operated for a period of 500 cycles and the predicted probabilities were recorded for every 5th cycle of operation. Finally, average of the predicted probabilities (depicted in figure \ref{fig:HeatMap}) were taken to test the model's consistency.


\subsection{Normal Operation of SV}
\begin{figure}[h]
  \centering
  \includegraphics[width=0.85\linewidth]{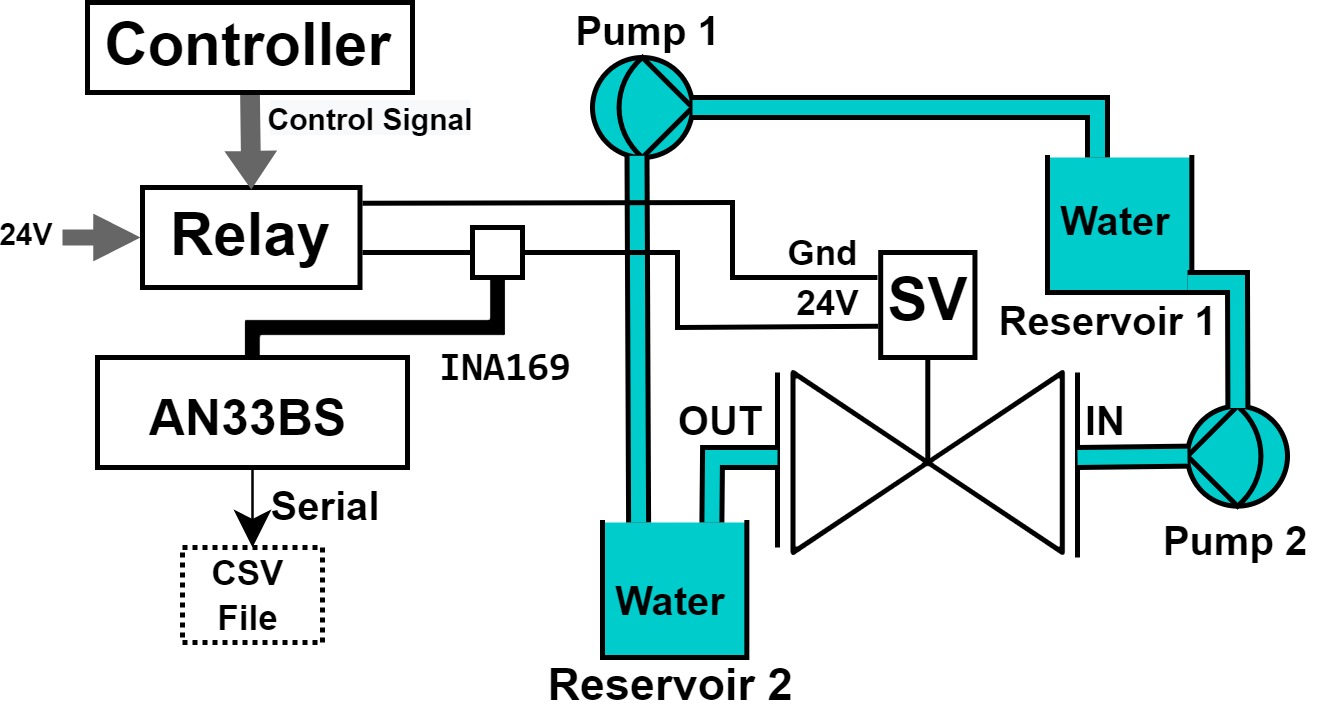}
  \caption{Normal Operation Block Diagrams}
  \label{fig:WaterLoad}
  \Description{}
\end{figure}

\begin{figure}[h]
  \centering
  \subfloat[di$/$dt vs Cycles for No load]{\includegraphics[width=0.5\linewidth]{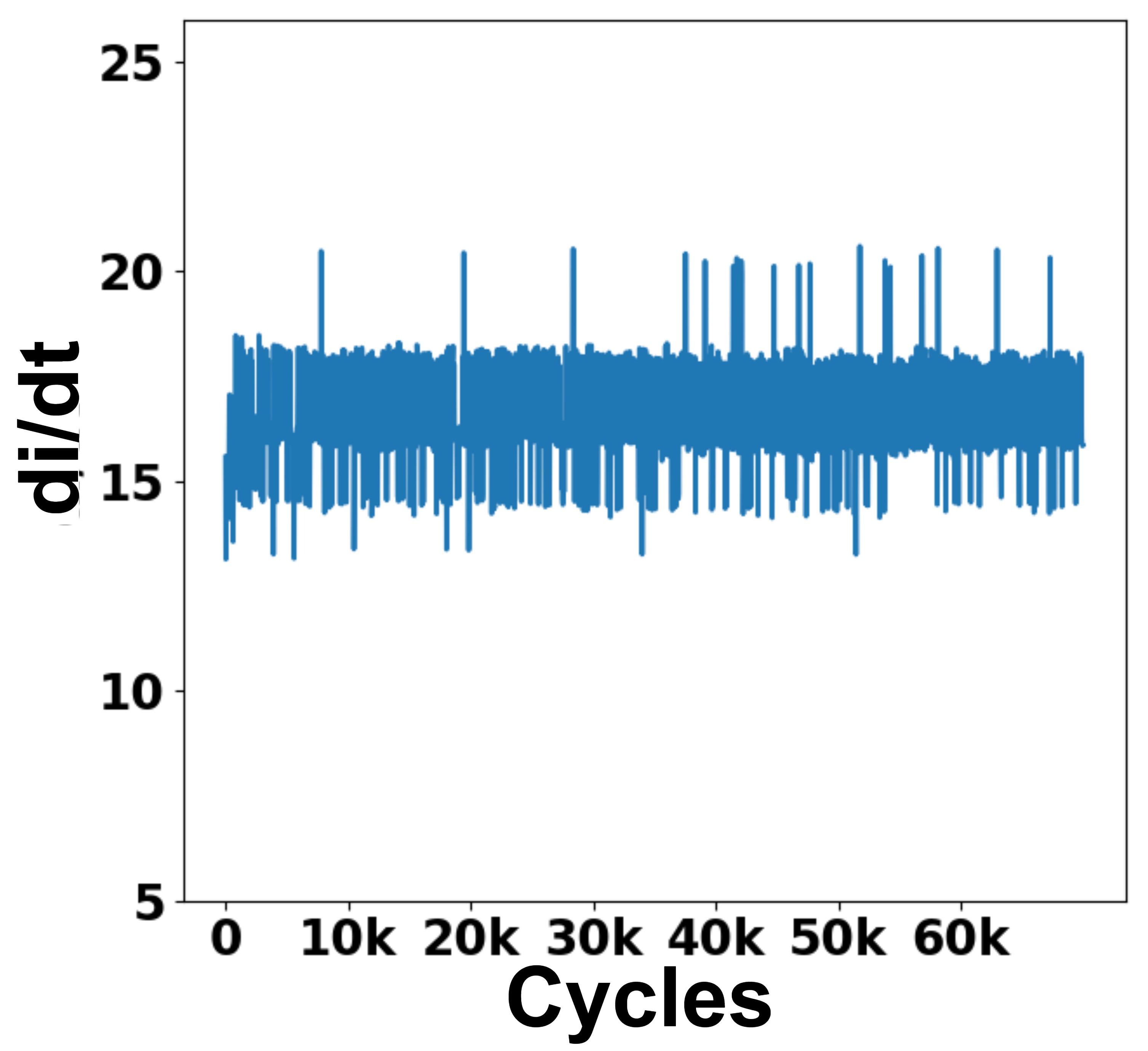}\label{fig:NoLoadResults}}
  \hfill
  \subfloat[di$/$dt vs Cycles with water as laod]{\includegraphics[width=0.5\linewidth]{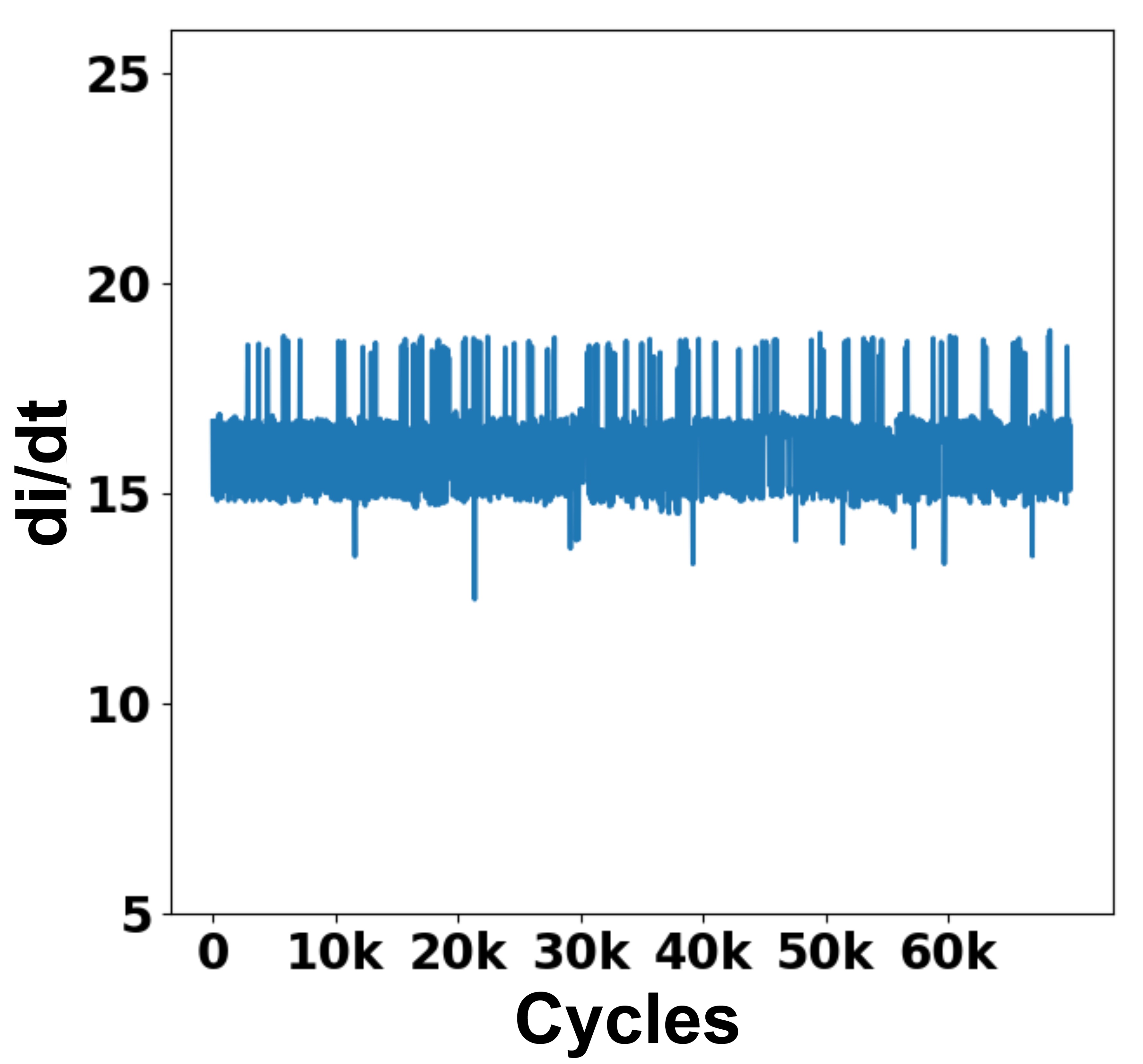}\label{fig:WaterLoadResults}}
  
  \caption{Feature Variation for the Two Scenarios}
  \label{fig:NormalOperation}
\end{figure}

Estimating the RUL of SV is analogous to predicting the remaining beats of the heart. SV exhibits a smooth functioning (as shown in figure \ref{fig:NormalOperation}) over a significantly long duration of time under normal conditions. For example, normal operations were noticed when we experimented with the SV in two different scenarios under it's operating conditions. In the first scenario, we studied SV in an open environment with no loading impact ( same setup as depicted in setup 1 of Figure \ref{fig:DataAcq} ). Whereas, in the second case, we carried out an experiment on SV employing water as the load. Figure \ref{fig:WaterLoad} illustrates the loaded condition test setup which consists of two water reservoirs one at higher altitude  (Reservoir 1) and one at lower altitude (Reservoir 2). Reservoir 1 is connected to the inlet of the SV through Pump 2. The use of the pump 2 is to make the water pound the valve at a higher pressure. The outlet of the SV is connected to reservoir 2. Pump 1 is utilized to pump water from reservoir 2 to reservoir 1 forming a perpetual loop.

Figure \ref{fig:NoLoadResults} and \ref{fig:WaterLoadResults} shows the results for the first scenario and second scenario respectively. In both the scenarios, SV was actuated at a rate of 0.5 Hz for a period of 2 days equivalent to 75000 cycles. Both the  scenarios conclude that there is no variation of di/dt indicating no change in state of the SV. In summary, normal operations are ensured until an unknown factor (x factor) triggers the SV to deteriorate.

\begin{figure*}[t]
  \centering
  \subfloat[Variation of Transient Response across for several cycles]{\includegraphics[width=0.47\linewidth,height=0.42\linewidth]{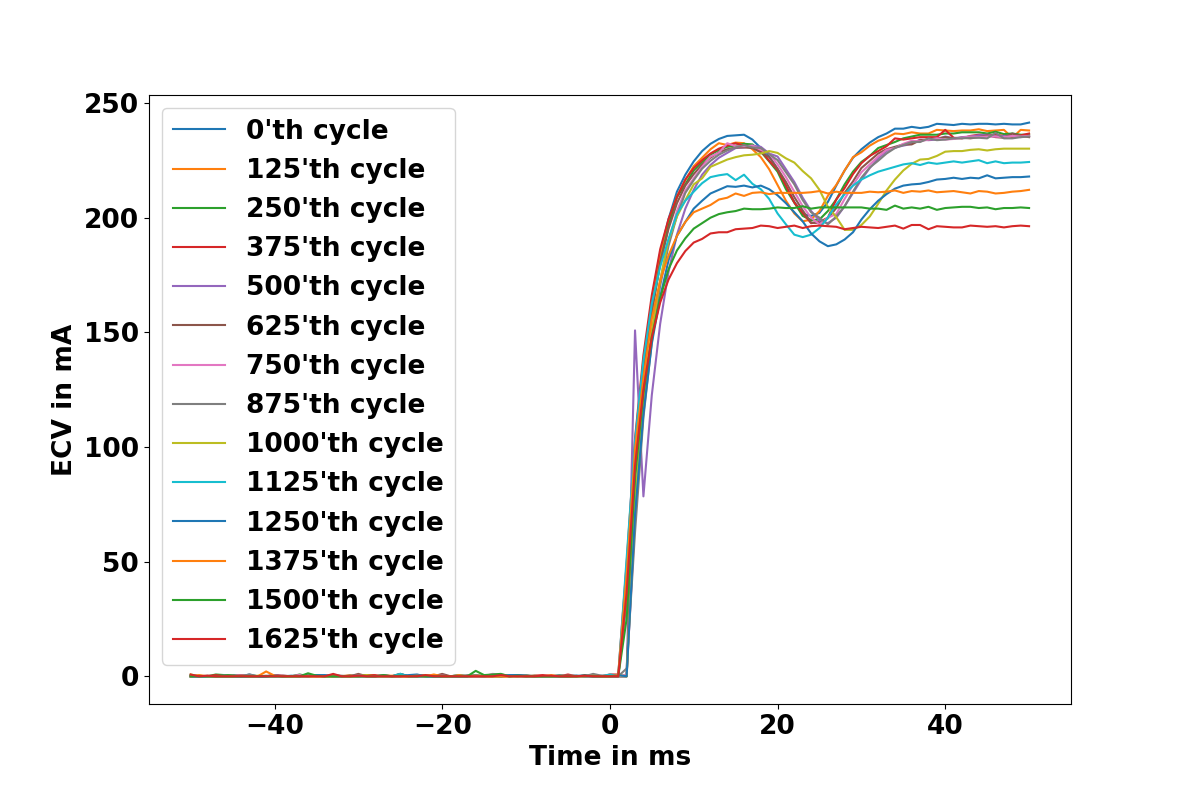}\label{fig:RULTransient}}
  \hfill
  \subfloat[Variation of AUC and di/dt]{\includegraphics[width=0.5\linewidth, height=0.37\linewidth]{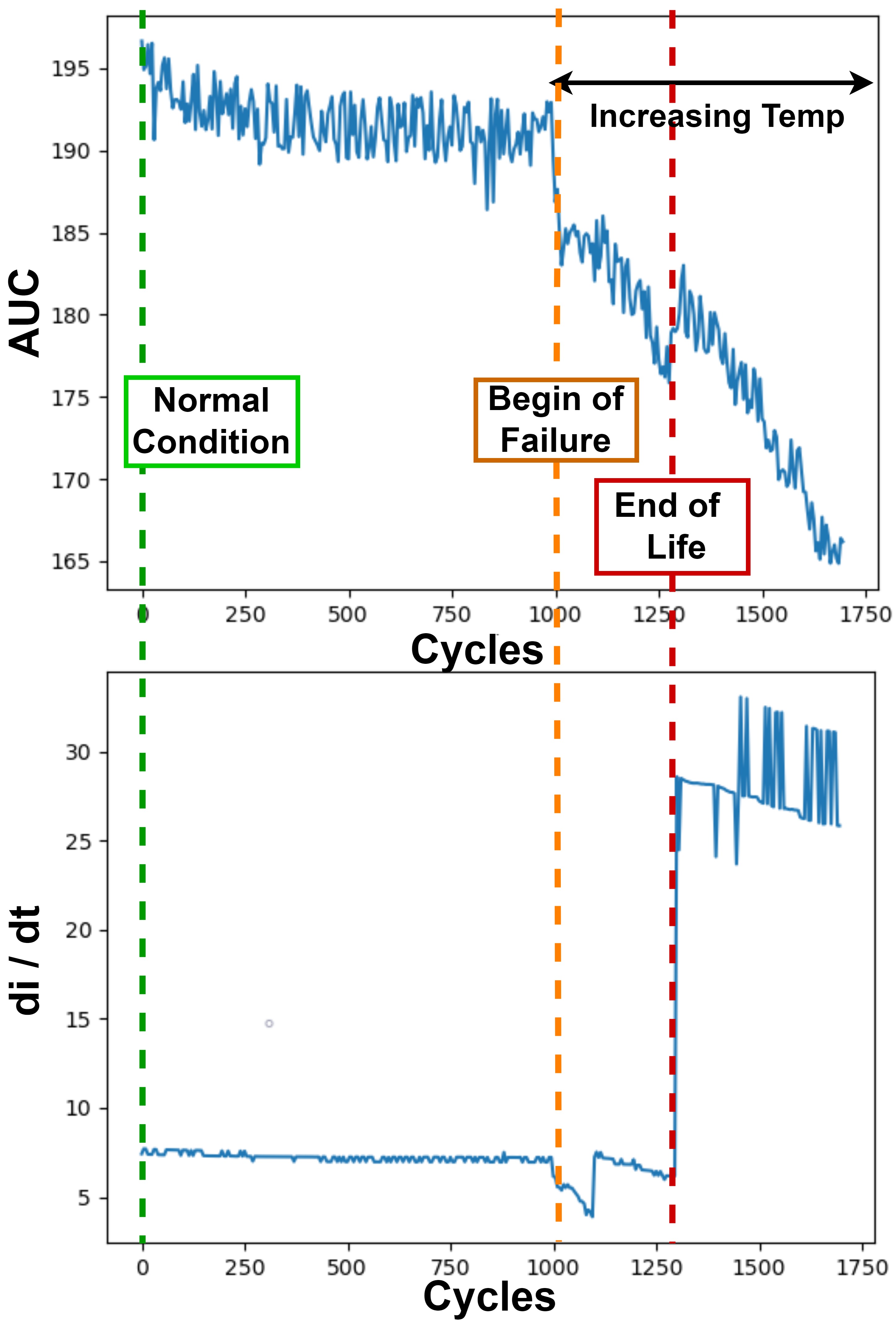}\label{fig:RULAuc}}\\
  
    \subfloat[RUL Loss Curve]{\includegraphics[width=0.26\linewidth]{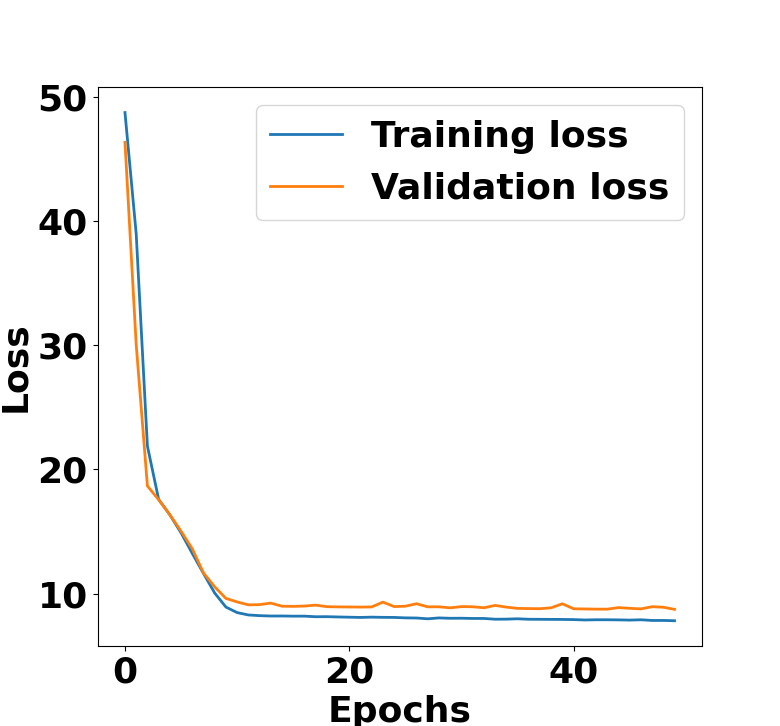}\label{fig:RULLossCurve}}
    \hfill
  \subfloat[Valve 1 Actual vs predicted RUL]{\includegraphics[width=0.36\linewidth]{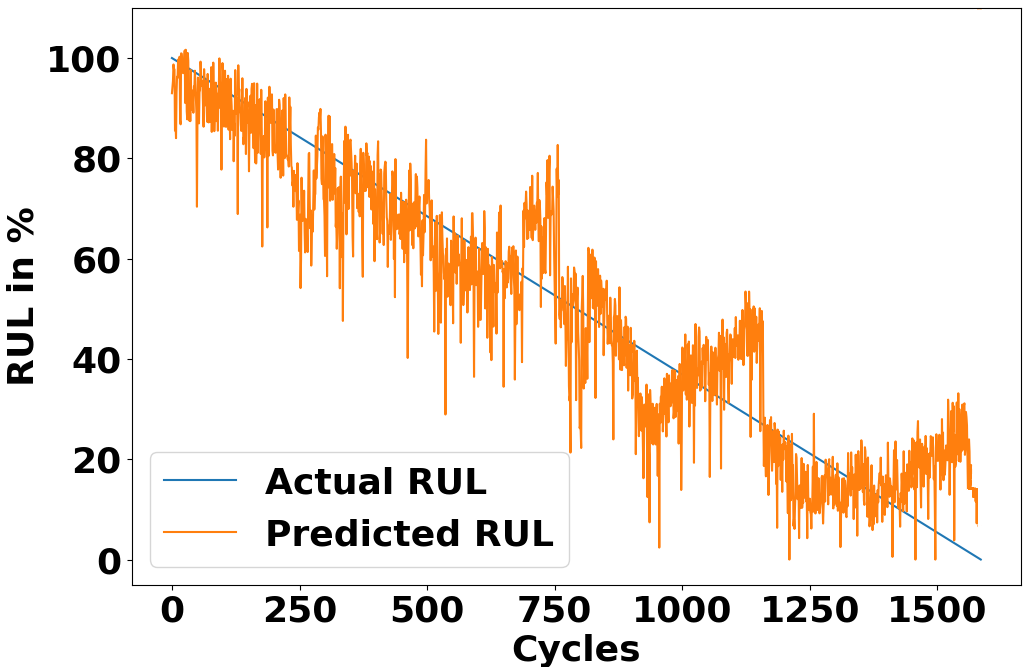}\label{fig:pvo1}}
    \hfill
  \subfloat[Valve 2 Actual vs predicted RUL]{\includegraphics[width=0.36\linewidth]{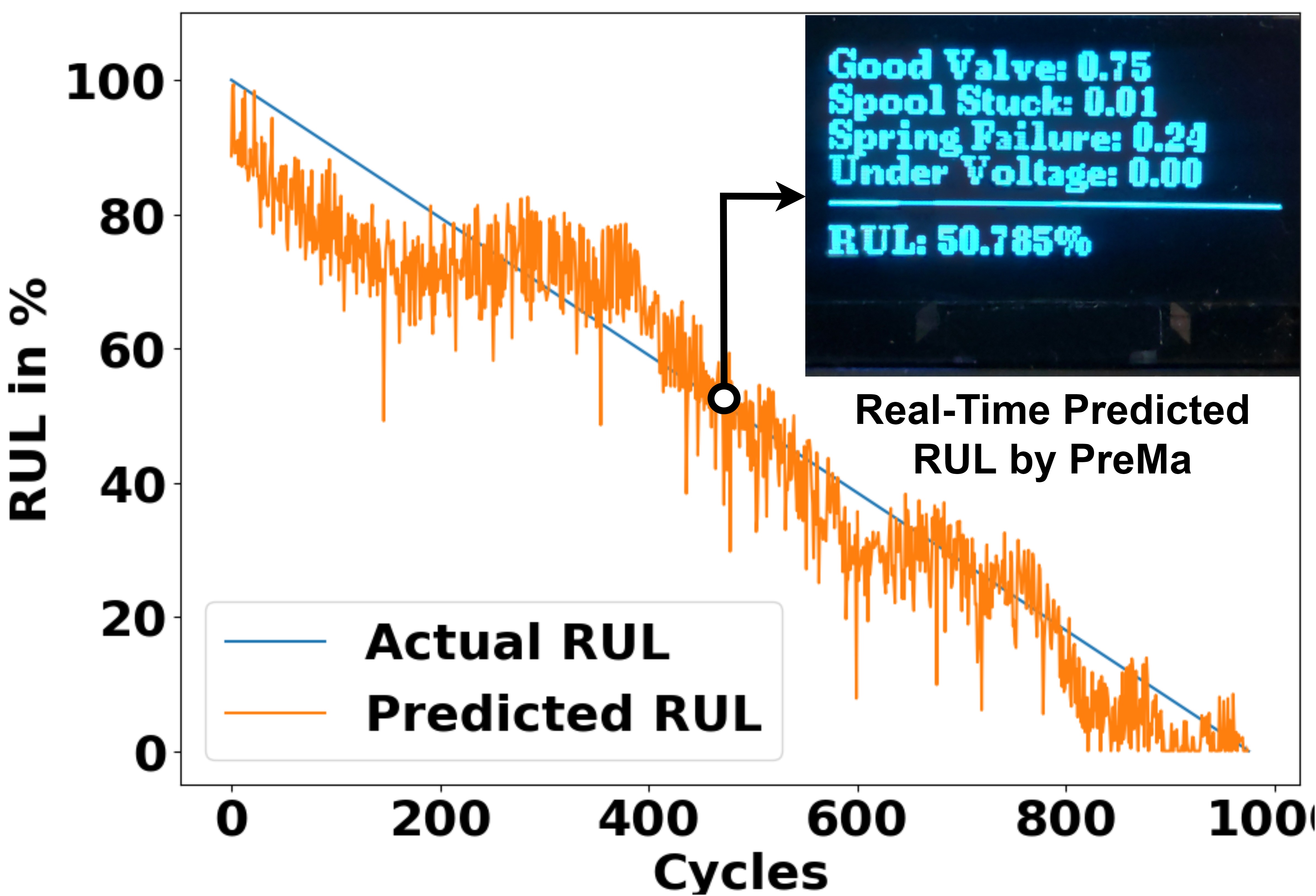}\label{fig:pvo2}}
    \caption{Variation of Transient Response, AUC and di/dt During Run-to-Failure Experiment}
\end{figure*}

\subsection{Triggering Condition for Start of Failure} \label{sec:TriggeringCondition}
Taking into account the above concerns, it is critical to find the triggering condition where SV starts deteriorating over time. The x factor to initiate this stage may arise dynamically due to any of the following circumstances in and around the SV's operating environment. The x factor could be: (a) Unregulated power supply to the SV caused by failure of switched-mode power supply, (b) Variation of plunger movement due to deposition of debris and (c) Coil heating. Many such scenarios can be speculated that might lead to commencement of failure. Section \ref{sec:RULExperementalSetup} explains the run to failure setup and conditions under which the SV's function deteriorates. On exposing the SV to higher temperature, a point of fault was noted at 70°C when there was no air at the outlet, implying that the SV failed to actuate. At this instance of time, our smart product is capable of classifying the SV's defect under the spool stuck fault class.

\subsection{RUL Results}
\label{sec:4.4}
\subsubsection{\textbf{Analysis during SV's life cycle}}
Figure \ref{fig:RULTransient} depicts the variation in transient response captured during the run to failure experiment described in section \ref{sec:RULExperementalSetup}. As we can witness in the figure \ref{fig:RULTransient}, there is a steady shift in transient response from good to faulty valve. At the end of experiment, we can observe that the dip seen in the healthy valve vanished and the resulting transient response resembles the spool stuck transient response of the valve. 

Figure \ref{fig:RULAuc} illustrates  (a) Normal operation (b) Start to failure and (c) End of life, meticulously captured by comparing cycle to cycle variation in the features that we extracted. The two prominent features served well for this purpose are di/dt and AUC. Figure \ref{fig:RULAuc} clearly indicates the variation of AUC during the valve degradation phase and di/dt indicates the point at which the valve fails to actuate completely; indicating the end of life. Considering these factors, both AUC and di/dt are used to train our deep neural regression model. The data set and the model used for training are described in section \ref{sec:NNRUL}.

\begin{figure*}[t]
  \centering
  \subfloat[1bar]{\includegraphics[width=0.2\linewidth]{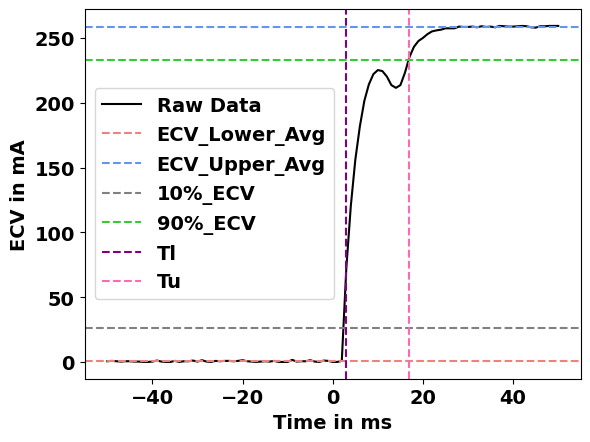}\label{fig:1bar}}
  \hfill
  \subfloat[2bar]{\includegraphics[width=0.2\linewidth]{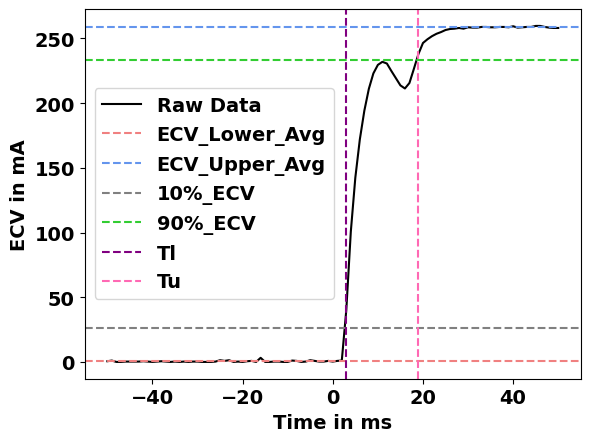}\label{fig:2bar}} 
    \hfill
  \subfloat[3bar]{\includegraphics[width=0.2\linewidth]{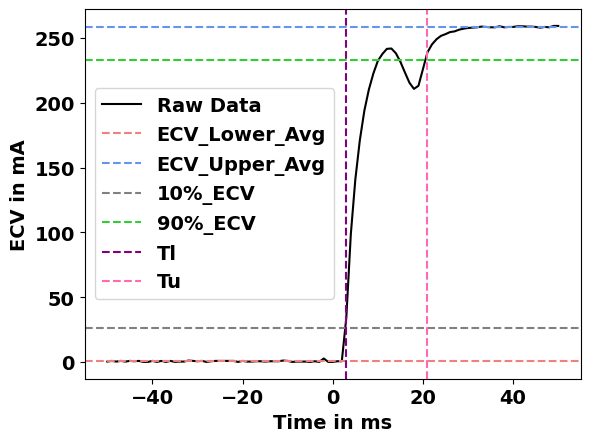}\label{fig:3bar}}
    \hfill
  \subfloat[4bar]{\includegraphics[width=0.2\linewidth]{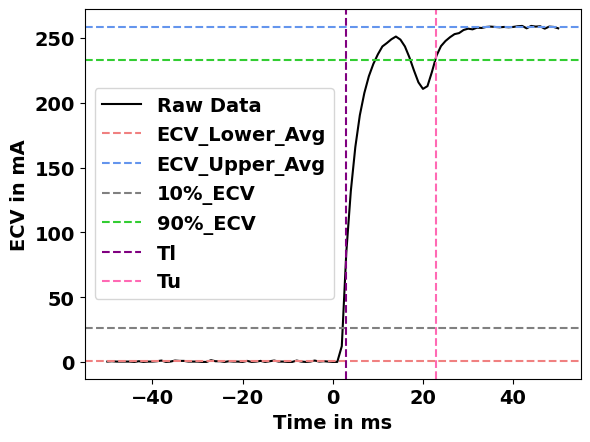}\label{fig:4bar}}
    \hfill
  \subfloat[5bar]{\includegraphics[width=0.2\linewidth]{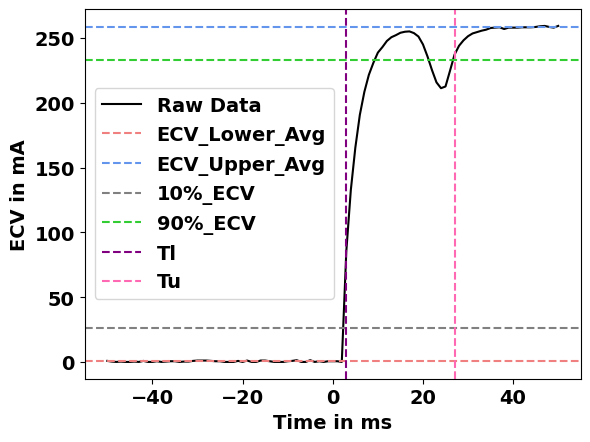}\label{fig:5bar}}
  
  \caption{Variation of Transient Response w.r.t Pressure}
\end{figure*}

\subsubsection{\textbf{On-device training and on-device inference}}

Similar to the section \ref{section:fd_train}, we tried out a combination of hyper parameters (epochs(10,50,100), batch size (5, 10, 30, 50) and optimizers (Adam, RMSProp, SGD)) for RUL NN regression model. Optimal loss (MAE) of 7.583 is obtained with the following parameters: batch size - 10, Optimizer - RMSProp and epochs - 50. Figure \ref{fig:RULLossCurve} is the loss curve which we obtained during training of the model. Figure \ref{fig:pvo1} and \ref{fig:pvo2} shows the actual vs predicted RUL for two different valves tested using the experimental setup as described in section \ref{sec:RULExperementalSetup}. Figure \ref{fig:pvo2} shows that our deployed PreMa can predict the RUL instantaneously and thus facilitating predictive maintenance.

The major takeaway from sections \ref{sec:FaultDetectionOfSV} to \ref{sec:4.4} leads us to importance of 'PreMa': real-time predictive maintenance in industrial operation of SV. It can be seen as an intelligent gadget that can detect the beginning of degradation caused by x factor(section \ref{sec:TriggeringCondition}) and then be used as a tool to estimate SV's remaining life and warn the industrial operator before the entire process is disrupted. Moreover, PreMa can also recognise the fault types caused due to the x factor. To conclude, our on-device inference facilitates both fault detection and RUL in real-time rolling out as a tiny embedded cognitive node.

\subsection{Effect of Environmental Condition}
\begin{figure}[h]
  \centering
  \includegraphics[width=0.9\linewidth]{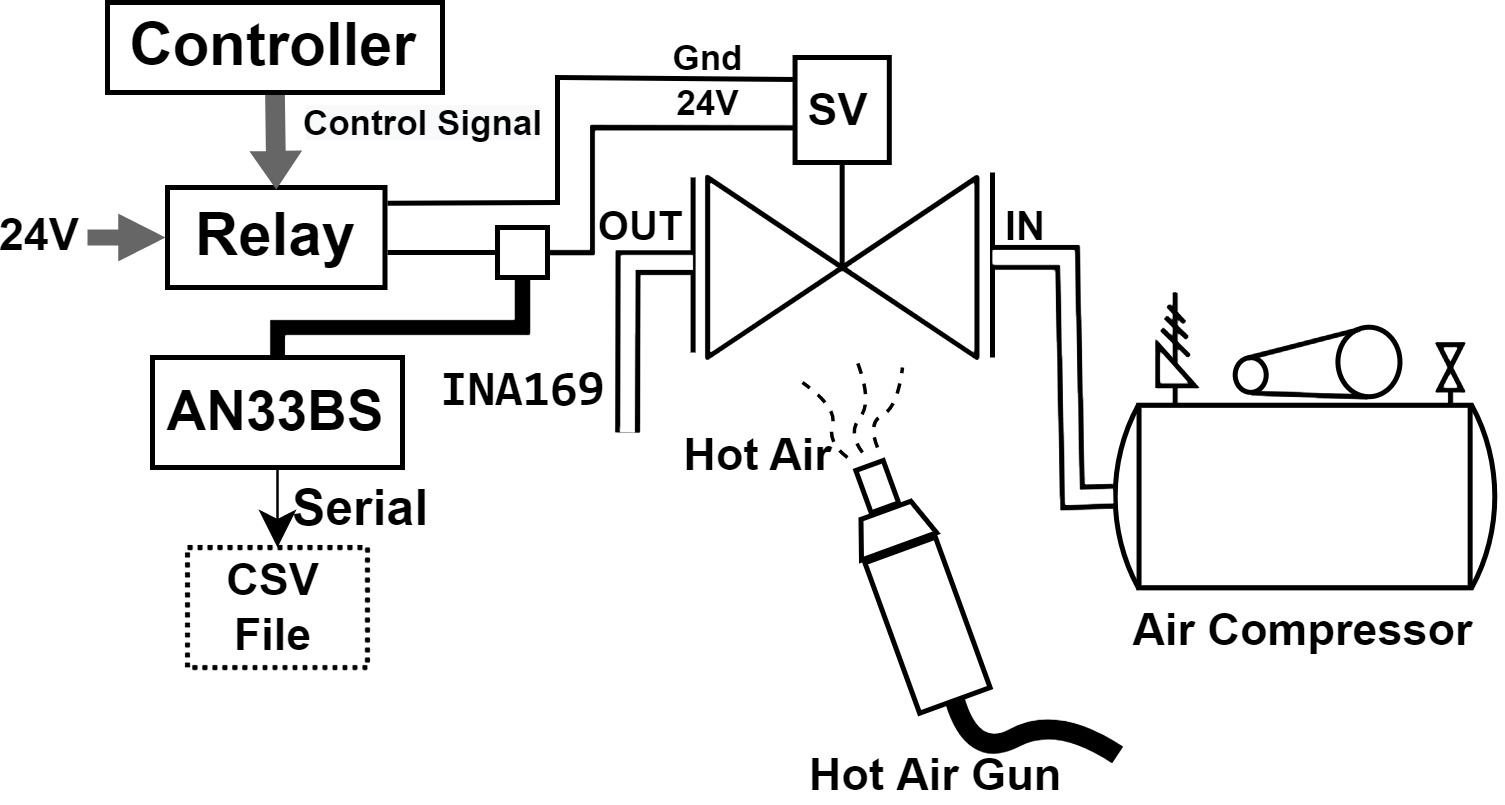}
  \caption{Environmental Condition Test Block Diagram}
  \label{fig:EnvironmentConditionBD}
  \Description{}
\end{figure}

Figure \ref{fig:EnvironmentConditionBD} shows experimental setup designed to test the impact of environmental factors such as pressure and temperature on SV. It comprises of SV with its control circuitry, our in-situ embedded hardware data acquisition system, an air compressor to vary the air pressure at the inlet of SV and a heat gun to emulate environmental temperatures. The experiment was conducted by varying the operating temperature of SV from $30^\circ$C to $100^\circ$C in increment of $10^\circ$C. For each temperature, 5 transient responses of SV were captured at pressure conditions ranging from 1 bar to 8 bar.


\subsubsection{\textbf{Effect of Temperature}}
\begin{figure}[h]
  \centering
  \includegraphics[width=0.8\linewidth]{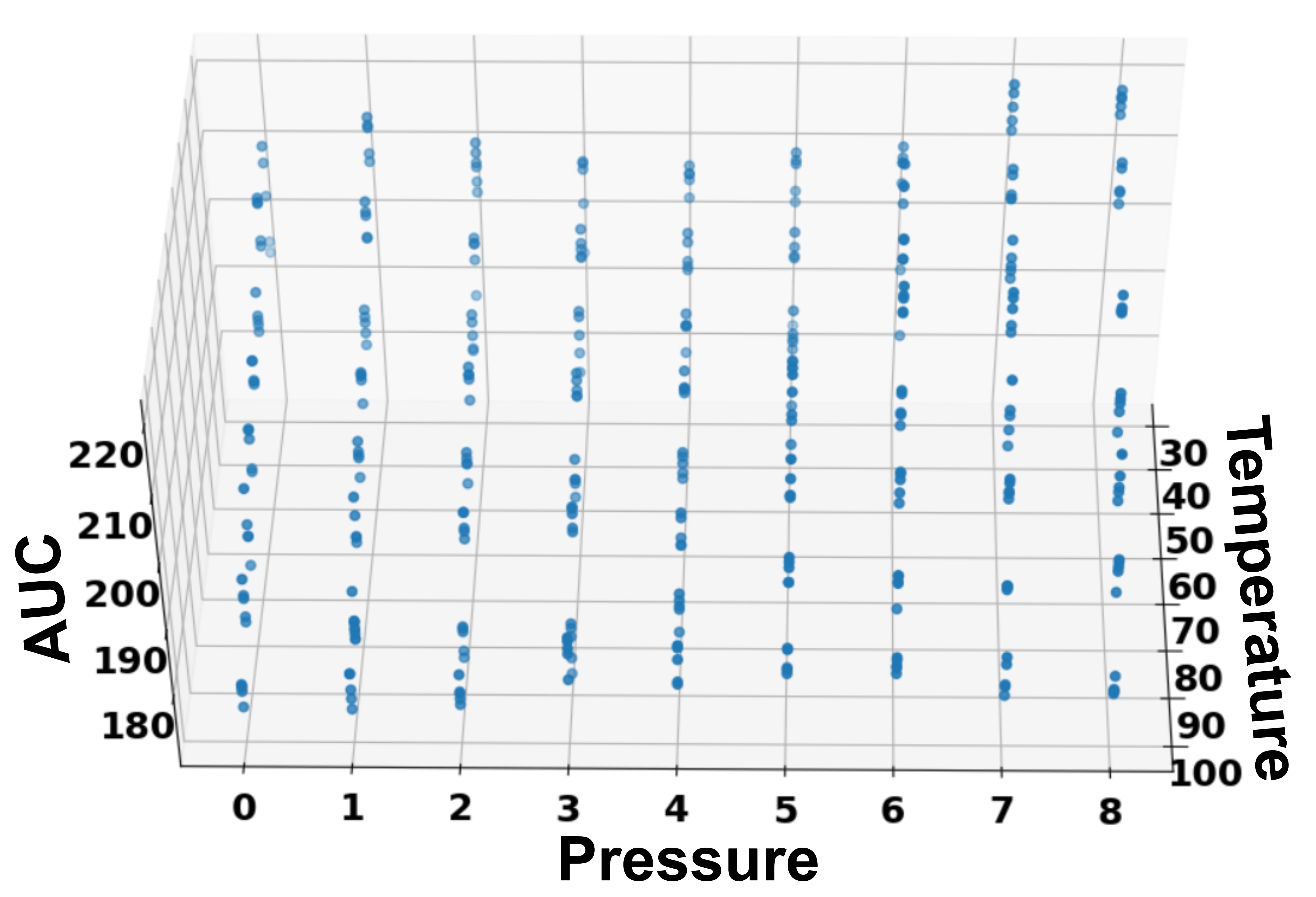}
  \caption{Effect of Temperature.}
  \label{fig:PTAUC}
  \Description{}
\end{figure}

Figure \ref{fig:PTAUC} depicts a three-dimensional plot of AUC, temperature, and pressure. By visualizing the graph, we can infer from the graph that for every instance of pressure (1bar - 8bar), the feature AUC decreases as temperature rises (30C - 100C).
\subsubsection{\textbf{Effect of Pressure}}
The effect of pressure on the SV's transient response is another important aspect we discovered. As seen over the sequence of figures from \ref{fig:1bar} to \ref{fig:5bar}, the peak of the transient response continues to rise upwards as pressure increases.

\subsection{Ping Pong Buffer Evaluation}
This section presents the tests carried out to evaluate the performance of ping pong buffer. Initially, three tests were performed by providing 100Hz, 10Hz and 1Hz sine wave signals with an amplitude of 2.4 V at the analog input of the AN33BS. These signals were sampled by the ADC at 1KHz. The reconstruction tests were validated and we observed that no data was lost during the buffer switching process as discussed in section \ref{sec:PingPongBuffer}. We further validated nrf52840 ADC's maximum sampling frequency of 200kHz by reconstructing a 20kHz input waveform.
The program was designed to allow dynamic change of sampling frequency and buffer size. Hence, we quantify `$C_{max}$' the maximum number of SV cycles, with frequency of operation `$f_{op}$', that can fit in a ping pong buffer of size `$K$' with sampling frequency set to `$f_s$' to enable fine tuning of these parameters to match industrial operational needs.\\

`$B_{fd}$' buffer fill duration, time taken to fill buffer of size $K$, with sampling frequency at $f_s$ is given by,
\begin{equation}\label{eq:5}
   B_{fd}=\frac{K}{f_s}\\
\end{equation}
The maximum number of SV cycles ($C_{max}$) the ping pong buffer can fit is dependent on  $B_{FD}$ and $f_{op}$ of SV actuation.
\begin{equation}\label{eq:6}
  B_{fd}=\frac{C_{max}}{f_{op}}\\
\end{equation}
Substituting equation \ref{eq:6} in \ref{eq:5}, 
\begin{displaymath}
  \frac{C_{max}}{f_{op}}=\frac{K}{f_s}\\ 
\end{displaymath}
Hence, $C_{max}$ is given by\\
\begin{equation}\label{eq:7}
 C_{max}=\frac{K\cdot f_{op}}{f_s}\\ 
\end{equation}

For Example, taking 
\begin{displaymath}
 K=10000,f_s=1kHz,f_{op}=0.5Hz\\
\end{displaymath}

After substituting in equation \ref{eq:7}
\begin{displaymath}
C_{max}=5Cycles
\end{displaymath}

From the above example, it can be deduced that a maximum of 5 SV actuation cycles with frequency of operation 0.5Hz can be stored in buffer 1 (size 10K) of the ping pong buffer, with sampling frequency of 1kHz. As discussed in section \ref{sec:PingPongBuffer}, current data pertaining to the initial 5 cycles are acquired into buffer 1 by easyDMA. Once  buffer 1 is full, it is sent forth to CPU for feature extraction followed by inference ( Fault detection, RUL estimation), concurrently the subsequent 5 cycles are stored in buffer 2 through easyDMA. When buffer 2 is full the acquired data is sent for inferencing, while buffer 1 gets overwritten with the subsequent 5 cycle data. This process keeps iterating. Thus, simultaneous operation of continuous sampling and inferring is maintained.

Table \ref{tab:SubProcessDuration} depicts $B_{fd}$ and $C_{max}$ for different values of $K$ and $f_{op}$. The sampling frequency $f_s$ is fixed at 1Khz. The table also shows `$IT_{pc}$', the inference time per SV actuation cycle and `$IT_{pb}$', the time taken to infer data present in one of the ping pong buffer of size $K$. Here $IT_{pc}$ depends on: time required for (a) Feature extraction. (b) Fault detection (c) RUL estimation for one SV actuation cycle. $IT_{pb}$ depends on: (a) $K$ ping pong buffer size , (b) $C_{max}$ maximum number of SV cycles in ping pong buffer of size $K$, (c) $IT_{pc}$ and (d) Other delay components.


\begin{table}[h]
\small
  \caption{$C_{max}$, $B_{fd}$, $IT_{pc}$ and $IT_{pb}$ for distinct values $K$ and $f_{op}$}
  \label{tab:SubProcessDuration}
  \begin{tabular}{p{0.05\linewidth} p{0.1\linewidth}  p{0.2\linewidth} p{0.07\linewidth} p{0.2\linewidth} p{0.1\linewidth} p{0.1\linewidth}}
    \toprule
    $K$ & $f_{op}$(Hz) & $B_{fd}$($\mu$s) & $C_{max}$ & $IT_{pc}$($\mu$s)& $IT_{pb}$($\mu$s)\\
    \midrule
     1k & 2   & 1000004 & 2   & 75322.65 & 195000.3\\
        & 1   & 1000003 & 1   & 75302.08 & 108184.9\\
    2k  & 2   & 2000003 & 4   & 75170.94 & 331397.4\\
        & 1   & 2000004 & 2   & 75270.33 & 212456\\
        & 0.5 & 2000003 & 1   & 75368.66 & 131035  \\
    5k  & 2   & 5000003 & 10  & 75281.39 & 889271.7 \\
        & 1   & 5000004 & 5   & 75138.58 & 456607   \\
        & 0.5 & 5000003 & 2.5 & 75309.61 & 317589.1 \\
    10k & 2   & 10000003 & 20 & 75415.47 & 1661852  \\
        & 1   & 10000004 & 10 & 75335.3  & 1043888  \\
        & 0.5 & 10000003 & 5  & 75286.74 & 636969.2 \\
  \bottomrule
\end{tabular}
\end{table}

It is evident from the table \ref{tab:SubProcessDuration} that $IT_{pb}$ is always lesser than $B_{fd}$ thus, ensuring concurrency between sampling and processing (feature extraction to inferencing) hence, preventing data loss.

\subsection{Performance Evaluation} \label{sec:PingPongBuffer}
Product performance evaluation is an effective way to ensure that the solution is at the forefront of the market. It analyses a set of parameters that the end customers would be assessing in order to determine the applicability of the solution. To this end, table \ref{tab:PerformanceEvalution} depicts the key parameters that provide a better insight about the industrial pertinence of the product.

\begin{table}[h]
\small
  \caption{Key Parameters of Product}
  \label{tab:PerformanceEvalution}
  \begin{tabular}{p{0.4\linewidth}  p{0.4\linewidth}}
  \toprule
    Parameters & Values\\
    \midrule
    Dimensions(L,B,H) & 8.4cm,7.0cm,3.74cm\\
    Volume & 219.912$cm^3$\\
    Weight & 105g\\
    Power Consumption & 140mW\\
    Latency ($IT_{PC}$) & 76.2ms\\
    Cost & 51.22 USD\\
  \bottomrule
\end{tabular}
\end{table}

As depicted in table \ref{tab:PerformanceEvalution}, the product is small in size, lightweight, responds quickly within 76.2ms without data loss and consumes 140mW of power, making it perfectly suitable for sustainable in-situ operations. Our product is an add-on and placed in series with the SV and easy to install. The program is efficient in terms of RAM and ROM utilization, and programmed for least user intervention and continuous operation.

\subsection{Generic Algorithm for Actuators }
As proposed in section \ref{section:intro}, our system can be extended to other electro-mechanical actuators, exhibiting similar types of transient responses during excitation. We considered an electro-mechanical relay as a similar actuator to leverage the scope of the proposed system. 

\begin{figure}[h]
  \centering
  \includegraphics[width=\linewidth]{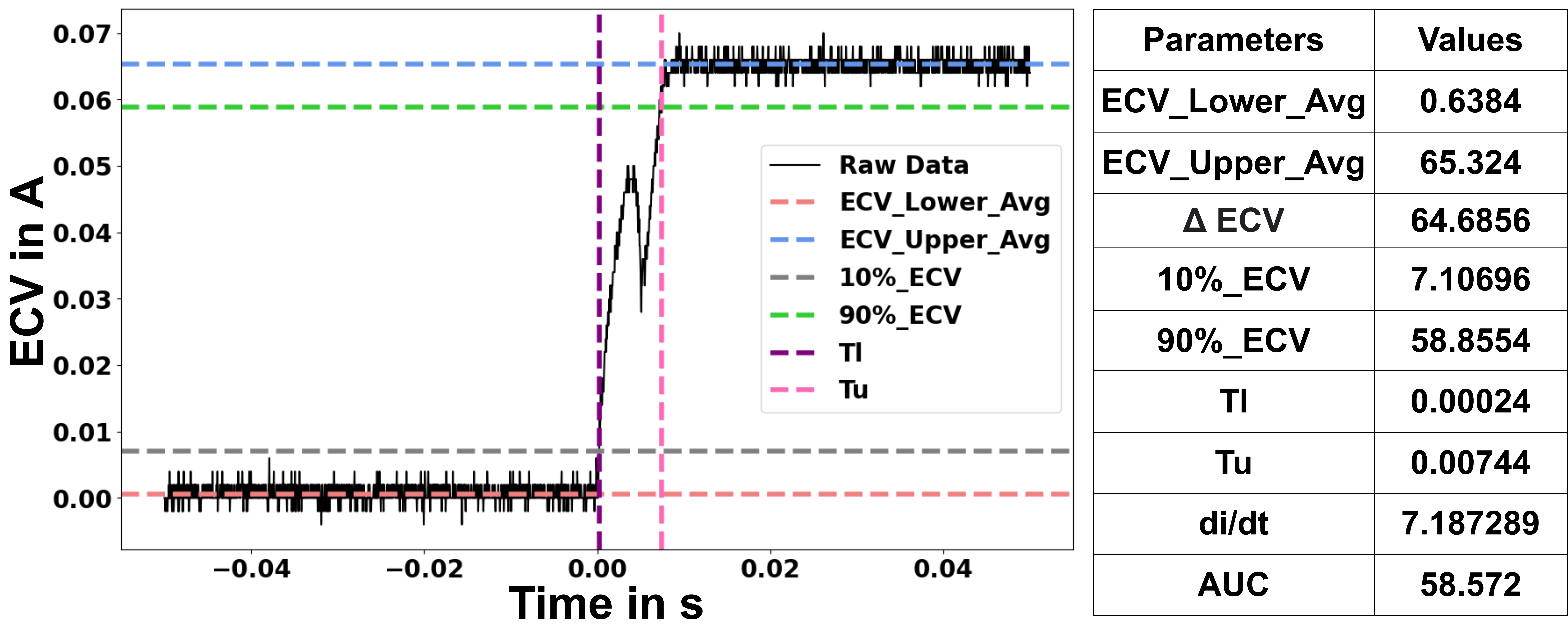}
  \caption{Feature Extraction for EM Relay}
  \label{fig:EMR}
  \Description{}
\end{figure}

Figure \ref{fig:EMR} depicts the features that we extracted for a relay via the same generic algorithm that we used for the predictive maintenance of the SV. The same end-to-end process that we followed in identifying faults and estimating the RUL of SV can be stretched out for health diagnosis of the relay.  

\section{Related Works}

\cite{[1]} is a product aimed towards efficient driving of SV during its steady state using PWM technique and incorporates plunger position detection circuit for fault detection but due to lack of edge intelligence and RUL estimation only reactive maintenance can be performed. \cite{[2]} is a product designed by Emerson capable of multiple types of fault detection analysing the pressure at inlet and out let of SV and incorporates RUL estimation. The use of multiple invasive sensors and the expensive nature of the product makes it less viable.

Taking into account the critical nature of SV, the importance of detecting its faults and estimating the RUL, many works have proposed evaluation methods for fault detection. \cite{[5]} Analyses the application of multi-kernel support vector machine with genetic optimization for fault classification and obtained a state-of-the-art accuracy of 98\%. Use of PCI9111A DAQ drives up the implementation cost. \cite{[6]} Puts forth a novel method for SV fault detection using vibration signal of SV. It uses a wavelet transom for noise reduction and a heuristic threshold function for Health Index calculation. \cite{[11]} Simulates the incorporation of an optical fiber between the spring and the plunger of SV in order to measure the force exerted by the plunger during its actuation and uses it to compares multiple standard ML algorithms for prediction of resistance. Lowest relative error of  0.03\% was obtained using gaussian process regression. \cite{[9]} Puts forward the ideology that coil resistance and the plunger clearance of SV in a diesel engine are the key indicators for degradation. It aims to deduce parameters correlating to the real world frictional forces acting up on the plunger of SV and trains an artificial NN for normal and fault classification. \cite{[6],[11],[9]} propose indirect methods of measurement and \cite{[6],[11]} suffer from damping issues which may alter the captured signals. \cite{[13]} Throws light on three types of fault that could occur in SV used in railway breaking system, that is, burn out of SV coil, debris accumulation and valve seat damage. It proposes two types of methods of fault detection: (a) the SV drive current based and (b) pressure based. The draw back of  this work is that it is a physics based model(discussed below) and with the use of licensed software (NI LabVIEW) there is a difficulty in scalability.

 RUL prediction is mainly classified into (a) model based approach and (b) data driven approach. Model based approach analyses the physics behind the failure mechanisms \cite{[10]}. However since SV is an amalgamation of magnetics, electronics and thermotics, this can encounter multiple types of faults during its lifetime and since its operation not only depends on internal and external environmental parameters, it is difficult to design an accurate physics based model. 
Whereas, data driven prognosis approach revolves around evolution of certain parameters through its lifetime.\cite{[3]} Uses AC SV current signatures for feature extraction. For fault detection, it compares the performance of support vector classification with other standard ML techniques and presents an accuracy of 94\%. It proposes a deep NN (DNN) regression algorithm for RUL estimation and compares its performance with XGBoost regressor and concludes that XGBoost performs better than DNN regression. The utilization of expensive DAQ(NI-9205) and licenced software (NI LabVIEW) makes the work less scalable. The use of Bayesian convolution NN in \cite{[7]} enables uncertainty estimation of SV current signature parameters. By using mean field variance, the model's mean absolute error was improved by nearly 60\%. The drawback of this work is that it uses powerful GPU enabled computers to train and perform prognosis driving the cost up. The works in \cite{[8]} proposes the application of an exponential degradation model for the deterioration of SV and a particle filter to mitigate the effect of noise. It dynamically combines the Bayesian updating and expectation maximization to compute the RUL. The proposed method obtained a state-of-the-art relative prediction error of 1.05\% to 2.56\%. \cite{[4]} The paper analyses the current response of multiple run-to-failure experiments for a SV. A degradation model is trained using Kalman filtering technique to estimate uncertainties and Monte Carlo sampling to generate future degradation samples. The predicted RUL was within 10\% of the actual RUL. Unlike our work, \cite{[8],[4]} does not take into account the effect of external conditions such as operating temperature and pressure of the medium at the inlet of SV.

\section{Conclusion}
This work proposes an intelligent embedded infrastructure with a low-cost data acquisition system, comparable with high end data acquisition system. Rather than extracting the extrinsic behaviour of the SV (vibration, temperature, etc) which is a function of the environmental conditions, we extract the internal behaviour, i.e. the drive current, by indigenously developing a sensing system for the same which acquires data concurrently by deploying ping pong buffer. This work also proposes a feature extraction module which has characteristics similar to a step response model of a dynamic model system. To bestow in-situ intelligence for the estimation of RUL and fault detection on the edge device, two NN models of 69 kb together were deployed on the AN33BS, yielding an accuracy of 96.45\% for fault detection and RUL loss of 7.583 (MAE). To make this infrastructure user friendly, the training of the model is implemented on mobile phone, facilitating its use by any industrial person. 

Hence, our product PreMa operates on low power of 140mW, has a low inference latency of 76.2ms, and comes at an economical price of 51.22\$. To conclude, PreMa is a compact cognitive edge solution for predictive maintenance in real-time and also a first of its kind generic infrastructure for numerous electro-mechanical actuators.


\section{Future Scope}

Although we have achieved a state of the art model in on-device training  on mobile phones and inferencing on microcontrollers in future both training and inference can be achieved  on  microcontrollers to confer the device with real-time learning characteristics. This feature will make the product more dynamic by tuning the device to suite the vivid application of SV hence, making the device more efficient. Moreover, it will mitigate the need of the mobile phones, making PreMa a better standalone product. Edge training can also lead to a new paradigm of federated learning (FL)[29]. FL can be integrated with PreMa using radio frequency communication forming a wide mesh network leading towards a robust internet of things and machine learning solution for SV health prognosis.
\bibliographystyle{ACM-Reference-Format}
\bibliography{sample-sigconf}

\appendix

\end{document}